\documentclass[review]{elsarticle}

\usepackage{lineno,hyperref}
\usepackage[ruled,linesnumbered]{algorithm2e}
\usepackage{booktabs}
\usepackage{graphicx}
\usepackage{amsmath}
\usepackage{amsfonts}
\usepackage{mathrsfs}
\usepackage{dsfont}
\usepackage{subfigure}
\usepackage{amsthm}

\usepackage{bbm}

\usepackage{setspace}
\usepackage{multirow} 
\usepackage{float}

\usepackage{color}
\usepackage[numbers]{natbib}

\journal{Journal of \LaTeX\ Templates}

\begin{document}

\begin{frontmatter}

\title{Topological Structure Learning for Weakly-Supervised Out-of-Distribution Detection}





\author[mymainaddress]{Rundong He\corref{co-author}}
\author[mymainaddress]{Rongxue Li\corref{co-author}}
\cortext[co-author]{Contribute to this work equally}
\author[mymainaddress]{Zhongyi Han\corref{mycorrespondingauthor}}
\author[mymainaddress]{Yilong Yin\corref{mycorrespondingauthor}}
\cortext[mycorrespondingauthor]{Corresponding authors}
\address[mymainaddress]{School of Software, Shandong University}

\begin{abstract}
Out-of-distribution~(OOD) detection is the key to deploying models safely in the open world. For OOD detection, collecting sufficient in-distribution~(ID) labeled data is usually more time-consuming and costly than unlabeled data. When ID labeled data is limited, the previous OOD detection methods are no longer superior due to their high dependence on the amount of ID labeled data. Based on limited ID labeled data and sufficient unlabeled data, we define a new setting called \underline{W}eakly-\underline{S}upervised \underline{O}ut-\underline{o}f-Distribution \underline{D}etection~(WSOOD). To solve the new problem, we propose an effective method called \underline{T}opological \underline{S}tructure \underline{L}earning~(TSL). Firstly, TSL uses a contrastive learning method to build the initial topological structure space for ID and OOD data. Secondly, TSL mines effective topological connections in the initial topological space. Finally, based on limited ID labeled data and mined topological connections, TSL reconstructs the topological structure in a new topological space to increase the separability of ID and OOD instances. Extensive studies on several representative datasets show that TSL remarkably outperforms the state-of-the-art, verifying the validity and robustness of our method in the new setting of WSOOD.
\end{abstract}


\end{frontmatter}

\section{Introduction}\label{sec:introduction}
The deep neural networks have achieved considerable success in the scenario where the training and testing data are sampled from an identical distribution \cite{canziani2016analysis, hao2020person}. However, in many real applications, the assumption cannot be satisfied due to the existence of unknowns. A reliable classification model ought to own the ability to say ``I do not know" to out-of-distribution~(OOD) data that the model has not seen before, which is the key to deploying models safely in the real world \cite{nguyen2015deep, yang2021generalized}. For example, a wildlife monitoring system with the ability to detect OOD data will not confidently regard unknown animal categories as known categories, which is essential to help humans discover new species \cite{nguyen2017animal}. In medical image recognition \cite{han2021semi, uwimana2021out}, models with the ability to detect OOD data can help doctors discover rare and novel diseases and prevent patients missing the best treatment period. In autonomous driving, OOD detection enables cars to evoke human control of driving in an emergency or unknown scenarios \cite{hell2021monitoring, nitsch2021out}, which contributes to safer and more reliable autonomous driving. 

OOD detection has received much attention because of its significance, and plenty of methods have emerged. The existing OOD detection methods can be divided into two main categories: classification-based OOD detection methods and density-based OOD detection methods. The classification-based methods contain post-hoc based methods \cite{hendrycks2016baseline, liang2017enhancing} and fine-tuning based methods \cite{hendrycks2018deep, liu2020energy}. \citet{hendrycks2016baseline} detected OOD data with the softmax confidence score. \citet{liang2017enhancing} used temperature scaling and input perturbation to amplify the OOD's separability. \citet{hendrycks2018deep} and \citet{liu2020energy} fine-tuned the model by introducing a large-scale auxiliary OOD dataset. Density-based methods in OOD detection explicitly generate the in-distribution~(ID) with some probabilistic models and regard test data in low-density regions as OOD data \cite{ren2019likelihood, xiao2020likelihood, zisselman2020deep}. Density-based methods are challenging to train and optimize, and the performance often lags behind the classification-based methods \cite{yang2021generalized}.

\begin{figure}[t]
\setlength{\abovecaptionskip}{0.cm}
	\centering
	\includegraphics[width=0.85\textwidth]{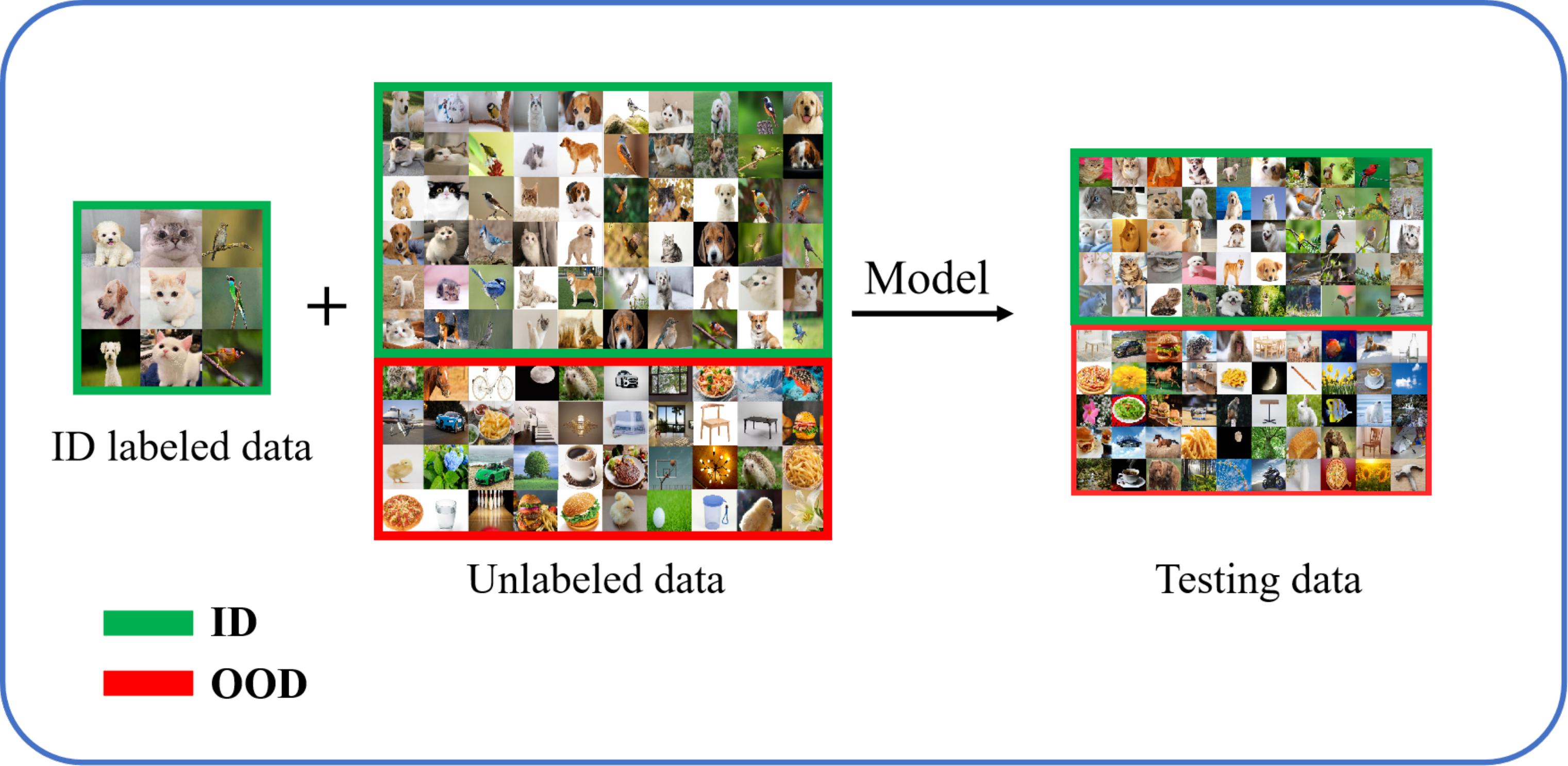}
	\caption{The setting of weakly-supervised OOD detection. While training, we utilize both limited labeled ID data and a large number of unlabeled data mixed with ID and OOD data. We do not know whether an example in the unlabeled dataset belongs to ID or OOD.
	}
	\label{fig1}
\end{figure}

Despite the excellent performance of OOD detection in the previous methods, they assume that ID labeled data is sufficient by default. However, this assumption cannot be satisfied in realistic scenarios because obtaining sufficient ID labeled data is usually costly and time-consuming. Especially in some specific fields, labeling data requires experienced experts. For example, it is expensive to hire an experienced doctor to label X-ray images in medical image recognition \cite{bustos2020padchest, dong2017learning, han2021semi}. When seriously lacking ID labeled data, the performance of the previous OOD detection methods declines severely, and it is hard to guarantee the security of the model. Accordingly, we refer to this problem as Weakly-Supervised Out-of-Distribution Detection~(WSOOD), which is shown in Fig.~\ref{fig1}. Solving this problem can contribute to deploying machine learning models more reliably in the real world.

In the WSOOD setting, limited ID labeled data and a large amount of unlabeled data are available. Compared to the previous settings of OOD detection, the key to WSOOD is that the amount of ID labeled data is limited, bringing three new challenges. Firstly, it is difficult for the model to learn the discriminable features of the ID and OOD. Most of the OOD detection models are trained with sufficient ID labeled data, allowing the model to learn ID reliable features. However, with limited ID labeled data, overfitting is likely to appear. Secondly, it is difficult to mine reliable positive and negative pairs from all the instances and effectively use them. If two instances form a positive pair, they are more similar and more likely to belong to the same class. Conversely, if two instances form a negative pair, they are very different and are less likely to belong to the same class. Due to difficult feature extraction, it is difficult to judge the connections between any two instances. Lastly, the differences between the ID and OOD instances are subtle. So far, there is only one specific method called STEP \cite{zhou2021step} to solve the problem of WSOOD. However, positive and negative pairs mined through STEP are not reliable enough. The mining process of positive pairs is too crude and straightforward, ignoring the fact that different positive pairs have different levels of closeness. STEP considers all other instances except itself as its negative instances, which is unreasonable and may be in conflict with positive pairs. Moreover, STEP does not fully exploit adequate information in labeled data.


To solve the existing problems and boost WSOOD, we propose \underline{T}opological \underline{S}tructure \underline{L}earning~(TSL). Firstly, TSL uses SimCLR \cite{chen2020simple} to obtain reliable feature representations on all labeled and unlabeled data, which help distinguish between ID and OOD and construct points for the topological space. Secondly, to enhance the reliability of mined pairs from all the data for the next topological structure reconstruction, TSL mines close positive pairs and loose positive pairs according to reciprocal and non-reciprocal neighbors. At the same time, TSL mines reliable negative pairs in the complement of positive pairs according to the Mahalanobis \cite{mclachlan1999mahalanobis} distance. Then, TSL reconstructs a new topological structure in a new topological space to increase the separability of ID and OOD. When constructing the new topological structure, TSL performs two steps.  The first is the maintenance of the topological skeleton. TSL maintains the topological skeleton by reducing the intra-class Mahalanobis distance of limited ID labeled data. The second is to extend and reconstruct the whole topological structure in the new topological space based on the topological skeleton and topological pairs mined before. TSL adjusts the distance between instances with different credibility depending on the types of topological structure pairs mined before, including close positive pairs, loose positive pairs, and negative pairs.

The following are the primary contributions of this work:
\begin{itemize}
\item
We investigate a new setting called weakly-supervised out-of-distribution detection, analyze why this problem is difficult, and propose a novel method called TSL.
\item
We mine close positive pairs and loose positive pairs according to reciprocal and non-reciprocal neighbors. And we mine
negative pairs in the complement of positive pairs according to the Mahalanobis distance.

\item

We propose to reconstruct the topological structure in a new topological space by constructing a topological skeleton and extending the topological skeleton to increase the separability of ID and OOD instances.


\item
We carry out extensive experiments on several benchmark datasets. Experimental results verify the effectiveness and robustness of TSL.
\end{itemize}

\section{Related Work}\label{sec:relatedWork}
In this section, we first introduce methods of OOD detection with post-hoc and OOD detection by fine-tuning, respectively, in Section~\ref{subsec:OOD Detection with Post-hoc} and Section~\ref{subsec:OOD Detection by Fine-tuning}. Then, we detail several existing methods of OOD detection with unlabeled data and point out the problem of these methods in Section~\ref{subsec:OOD Detection with Unlabeled Data}. Finally, we introduce weakly supervised learning in Section~\ref{subsec:weakly-supervised learning}.

\subsection{OOD Detection with Post-hoc}\label{subsec:OOD Detection with Post-hoc}
Post-hoc based methods have the advantage of being easy to use without modifying the training procedure and objective \cite{yang2021generalized}. \citet{hendrycks2016baseline} used the maximum softmax value to detect ID and OOD data and regarded instances with the lower maximum softmax value as OOD data. \citet{liang2017enhancing} used temperature scaling and input perturbation to amplify the ID/OOD separability. \citet{liu2020energy} replaced the maximum softmax with a theoretically unbiased energy function and loaded it after any trained classifier. \citet{wang2021can}  further proposed JointEnergy score. \citet{lee2018simple} presented a score using the Mahalanobis distance concerning the closest class-conditional distribution in the feature space of the pre-trained network, which verifies that information of feature space also contributes to OOD detection. \citet{morteza2021provable} derived analytically an optimal form of OOD scoring function called GEM (Gaussian mixture based Energy Measurement), which is provably aligned with the true log-likelihood for OOD detection. However, post-hoc based methods still assign high-confidence predictions to OOD data due to the lack of supervision signals from OOD data \cite{du2022vos}.

\subsection{OOD Detection by Fine-tuning}\label{subsec:OOD Detection by Fine-tuning}
Fine-tuning based methods are characterized by utilizing extra data to enhance the OOD detection performance. According to the data source, fine-tuning based methods can be categorized into three branches: using auxiliary OOD datasets, using generated virtual outliers, and using unlabeled data. In the branch of utilizing the auxiliary OOD datasets, representative works include OE \cite{hendrycks2018deep}, ATOM\cite{chen2021atom}, Energy \cite{liu2020energy}. OE \cite{hendrycks2018deep} fine-tunes the classifiers that predict ID categories by using large-scale, complex, and diverse auxiliary OOD datasets, enabling models to transfer the performance of OOD detection to unseen OOD instances heuristically. By circular mining, ATOM \cite{chen2021atom} mines OOD data in the auxiliary OOD datasets according to the closeness with ID boundaries. Based on outlier mining framework, POEM \cite{ming2022poem} mines better the boundary between ID and OOD data by posterior sampling. Energy \cite{liu2020energy} introduces large-scale auxiliary OOD datasets for fine-tuning to create the energy gap between ID and OOD data. However, such approaches using an auxiliary OOD dataset rely heavily on the scale and diversity of the auxiliary OOD dataset, and obtaining a completely clean large-scale auxiliary OOD dataset is challenging \cite{shafaei2018less, yang2021generalized}. Another branch is to synthesize OOD data, such as VOS \cite{du2022vos}, OOD-MAML \cite{jeong2020ood}, CSI \cite{tack2020csi}. VOS \cite{du2022vos} extracts virtual outliers from feature space's low probability portion of the estimated class-conditional distribution. By gradient updating for particular meta-parameters, OOD-MAML \cite{jeong2020ood} constructs adversarial OOD instances. CSI \cite{tack2020csi} suggests generating virtual OOD data by hard or distribution-shifting augmentations. Nevertheless, the OOD signals offered by synthesizing virtual instances are not reliable and difficult to generalize to OOD test data. Another branch of fine-tuning-based OOD detection methods is utilizing unlabeled data, which will be introduced in detail in Section~\ref{subsec:OOD Detection with Unlabeled Data}.

\subsection{OOD Detection with Unlabeled Data}\label{subsec:OOD Detection with Unlabeled Data}
Fine-tuning based OOD detection methods with unlabeled data boost the performance of OOD detection by introducing unlabeled data. The unlabeled and test data are sampled from an identical distribution that contains both ID and OOD classes. Large-scale unlabeled data is convenient and inexpensive to obtain. ID and OOD signals embedded in unlabeled data can mitigate the concerns of limited ID labeled data and contribute to amplifying ID/OOD separability. The existing OOD detection methods with unlabeled data include UOOD \cite{yu2019unsupervised}, SCOOD \cite{yang2021semantically}, STEP \cite{zhou2021step}, GBND \cite{sun2021gradient}, etc. UOOD \cite{yu2019unsupervised} uses unlabeled data to maximize the discrepancy between the decision boundaries of the two classifiers after the public feature extractor and push OOD data outside of ID decision boundary. The key to SCOOD \cite{yang2021semantically} is to select ID instances from unlabeled data by unsupervised clustering of ID labeled data and unlabeled data. STEP \cite{zhou2021step} obtains reliable features of ID labeled data and unlabeled data through a simple contrastive learning method, SimCLR. It then mines positive and negative pairs for contrast learning, thus keeping OOD instances away from ID instances and obtaining good OOD detection performance. GBND \cite{sun2021gradient} constructs a gradient-based OOD detector, extracting OOD instances from unlabeled instances using the Mahalanobis distance. GBND obtains good OOD performance by using better and better OOD signals in the process of selecting OOD instances over and over again. \cite{katz2022training} proposes a framework for constrained optimization on unlabeled wild data mixed by ID and OOD classes and applies ALM (Augmented Lagrangian Method) on a deep neural network to solve this constrained optimization problem. Although these methods using unlabeled data are closer to real-world situations than those using auxiliary OOD datasets and get more OOD signals than those generating virtual instances, these methods are still based on a strong assumption that ID labeled data is sufficient. This assumption is difficult to be satisfied in several real-world applications because of costly labeling. Therefore, the new problem of out-of-distribution detection with limited in-distribution labeled data needs studying urgently.

\subsection{Weakly Supervised Learning}\label{subsec:weakly-supervised learning}

Weakly supervised learning aims to build models by weaker supervised signals to save data labeling costs. There are three main types of weakly supervised learning \cite{zhou2018brief,li2019towards}: inexact supervision, inaccurate supervision and incomplete supervision. Inexact supervision is characterized by only coarse-grained labels are given while training \cite{carbonneau2018multiple}. Inaccurate supervision is that the labels given while training is not necessarily true \cite{frenay2013classification}. Incomplete supervision is that only a small portion of the data in the training set is labeled, and a large amount of the remaining data is unlabeled \cite{chapelle2009semi,lee2003learning}.

Positive-Unlabeled~(PU) learning \cite{lee2003learning} is a case of incomplete supervised learning \cite{zhou2018brief} to deal with classification problems in the case of only positive instances and unlabeled data. Existing approaches to PU learning can be broadly classified into three categories. The first is two-step techniques \cite{liu2002partially, zhang2018anomaly, he2021robust, he2022towards}, which select reliable positive and negative instances from unlabeled data, and then use the supervised signals from original positive instances, reliable positive and negative instances selected before to widen the gap between positive and negative samples. ADOA \cite{zhang2018anomaly} is one of the two-stage methods for PU learning. The second category is biased learning \cite{liu2003building, han2021semi}, where the unlabeled data is treated as negative instances. The third is the class prior estimation methods \cite{lee2003learning} assessing the class prior to positive classes. If we use methods of PU learning for weakly-supervised OOD detection, we consider ID and OOD data as positive and negative instances, respectively. However, a default assumption of PU learning is that the proportion of positive instances in unlabeled data is relatively small, which is quite different from the problem we study. However, in our setting, the ratio of positive instances is unconstrained, so the methods of PU learning are not suitable to solve our problem.

\section{Method}\label{sec:method}

\begin{figure*}[tb]
\centering
\includegraphics[width = 1\textwidth]{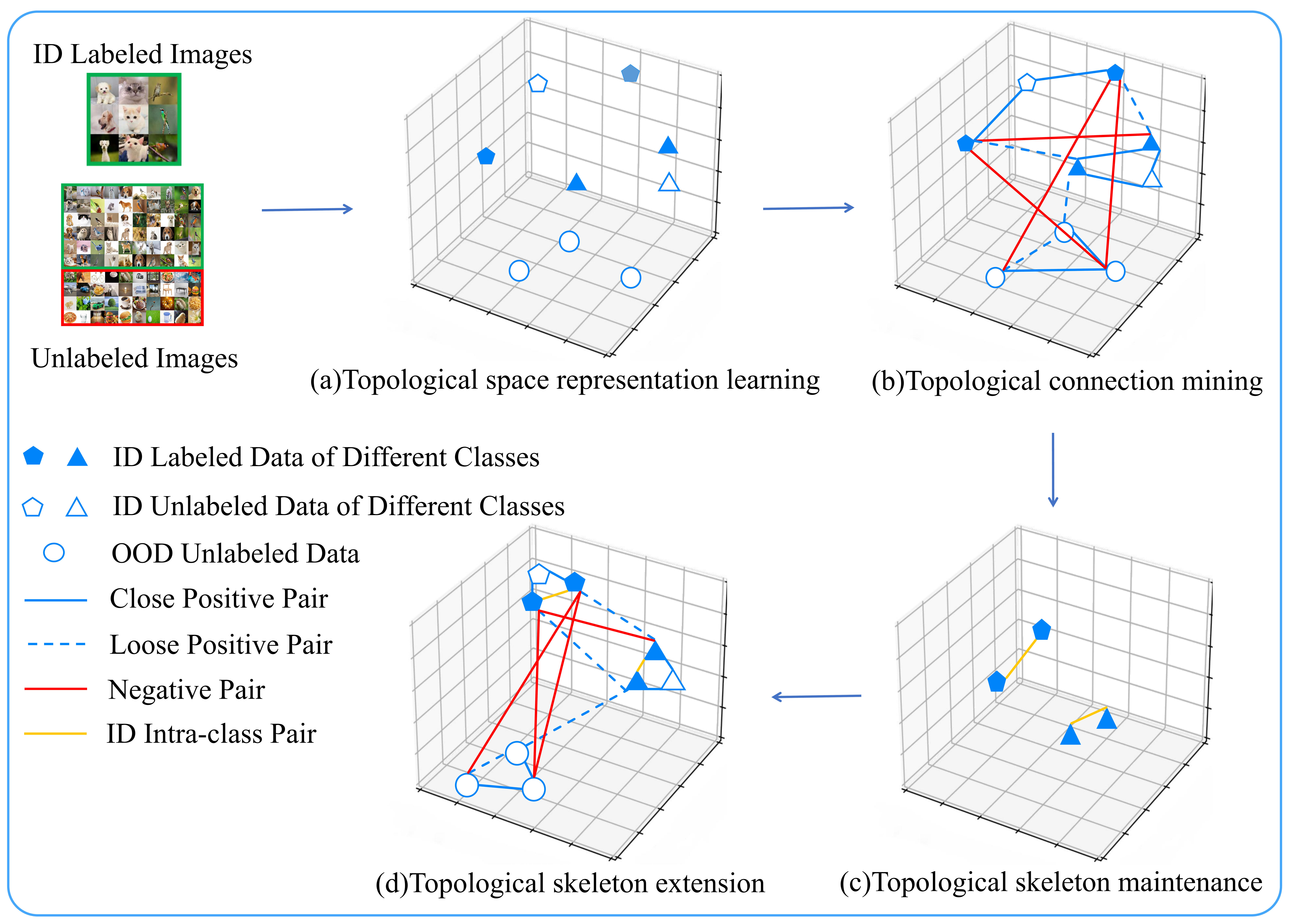}

\caption{The overview of TSL for WSOOD. (a) denotes that TSL obtains reliable features of ID labeled data and unlabeled data with a simple contrastive learning method, SimCLR. (b) denotes that TSL mines topological connections, including close positive pairs, loose positive pairs, and negative pairs. (c) denotes the topological skeleton maintenance with ID labeled data in the new topological space. (d) denotes topological skeleton extension with mined positive and negative pairs in the new topological space. Under the joint work of (c) and (d), TSL constructs a new topological structure in a new topological space.}
\label{fig2}
\end{figure*}


In this section, we elaborate on how our method, \underline{T}opological \underline{S}tructure \underline{L}earning~(TSL), works. Firstly, we describe the problem statements in Section~\ref{sec:Problem Statement}. Secondly, we extract reliable feature representations with SimCLR to construct the initial points for the topological space in Section~\ref{sec:Preliminary}. Then, to construct edges for the topological space and make preparation for the next topological structure reconstruction, we propose topological connection mining in Section~\ref{sec:Positive Negative Pair Mining}. Next, to increase the separability of ID and OOD instances, we reconstruct the topological structure in a new topological space in Section~\ref{sec:Topological Space Reconstruction}. When reconstructing the topological structure, TSL performs topological skeleton maintenance in the new topological space in Section~\ref{sec:Topological Skeleton Maintenance} and extends the whole topological structure based on topological skeleton with topological connections mined before in Section~\ref{sec:Topological Skeleton Extension}. What’s more, we summarize TSL’s optimizing objective in Section~\ref{sec:Optimizing Objective}. Finally, we introduce the scoring function of TSL in Section~\ref{sec:Scoring Function Designing}. The overview of TSL can be seen in Fig.~\ref{fig2}.

\subsection{Problem Statement}\label{sec:Problem Statement}
In this part, first of all, we introduce common OOD detection. Then, we introduce OOD Detection with sufficient unlabeled data consisting of ID and OOD data. After that, we introduce the newly proposed weakly-supervised OOD detection, provide formal descriptions of the new scenario and introduce notations used throughout this paper. Important notations and descriptions can be found in Table~\ref{tab0}.

\begin{table}[h]
   
\caption{The notations and their descriptions.}

\begin{tabular}{c|l}
\toprule
Notation & Description                          \\ \midrule
$D_{L}$       & ID labeled data                      \\
$D_{U}$       & Unlabeled data set                   \\
$D_{U}^{in}$       & Unlabeled ID data set                   \\
$D_{U}^{out}$       & Unlabeled OOD data set                   \\
 $\ D_{test}$    & Test set                             \\
$\ D_{test}^{in}$  & ID test set                          \\
$\ D_{test}^{out}$ & OOD test set                         \\
$\mathcal{X}$        & Feature space                        \\
$\mathcal{Y}$       & Label space                          \\
$P(\mathcal{X}, \mathcal{Y})$   & ID joint distribution                \\
$P(\mathcal{X})$     & ID margin distribution               \\
$Q(\mathcal{X})$     & OOD margin distribution              \\
$G(\mathrm{\boldsymbol{x}})$     & OOD decision function                \\
$k$  & Number of ID classes\\
$K$     &  Number of neighbors for the KNN algorithm\\
$n$      &   The number of labeled ID data\\
$m$    &   The number of unlabeled data\\
$\mathbf{P}$        & Projector for training               \\
$f$        & Feature extractor               \\
$\beta$     & Coefficient of negative pairs mining \\
$M$ & Margin for a loss function\\
$\lambda_1$    & Coefficient of topological skeleton maintenance         \\
$\lambda_2$    &Coefficient of close positive pairs          \\
$\lambda_3$    &Coefficient of loose positive pairs          \\

 \bottomrule

\end{tabular}
 \label{tab0} 
\end{table}

\paragraph{OOD Detection} While training, the available information is a labeled set $D_{L}=\left\{\left(\mathbf{x}_{i}, y_{i}\right)\right\}_{i=1}^{n}$ from the ID joint distribution $P(\mathcal{X}, \mathcal{Y})$, where $\mathbf{x}_{i}\in \mathcal{X}$, $y_i \in \mathcal{Y}$, $n$ denotes the number of ID labeled instances, $\mathcal{X}$ denotes the feature space, and $\mathcal{Y}$ denotes the label space. $\mathcal{Y} = \left\{1,\cdots, k\right\}$, where $k$ denotes the number of ID classes. Let $\ D_{test}$ denotes the test set, which consists of ID test set $\ D_{test}^{in}$ from ID marginal distribution $P(\mathcal{X})$ and OOD test set $\ D_{test}^{out}$ from OOD marginal distribution $Q(\mathcal{X})$, $P(\mathcal{X})\cap Q(\mathcal{X}) = \emptyset$. The goal of OOD detection is to obtain a decision function $G$ such that for a given test input $\mathbf{x} \in \ D_{test}$,
\begin{equation}
G(\mathrm{\mathbf{x}})= \begin{cases}0 & \text { if } \mathbf{x} \sim Q(\mathcal{X})\,, \\ 1 & \text { if } \mathbf{x} \sim P(\mathcal{X})\,,\end{cases}
\label{G}
\end{equation}
where $G(\mathbf{x})=1$ means that $\mathbf{x}$ belongs to ID data and $G(\mathbf{x})=0$ means that $\mathbf{x}$ belongs to OOD data.

\paragraph{OOD Detection with Sufficient Unlabeled Data} Different from the common OOD detection methods, in this setting, large-scale unlabeled data is available in addition to $D_{L}$. Let $D_{U}=\left\{\mathbf{x}_{i}\right\}_{i=1}^{m}$ denotes unlabeled data set mixed by both ID and OOD unlabeled data, where $m$ denotes the number of unlabeled data. In $D_{U}$, we denote the set of data from $P(\mathcal{X})$ by $D_{U}^{in}$ and the set of data from $Q(\mathcal{X})$ by $D_{U}^{out}$.

\paragraph{\underline{W}eakly-\underline{S}upervised \underline{O}ut-\underline{o}f-Distribution \underline{D}etection~(WSOOD)} The previous OOD detection methods assume that ID labeled data is sufficient, which is hard to satisfy in many real-life applications. However, obtaining unlabeled data is usually cheap. Then we define limited ID labeled set by $D_{L}=\left\{\left(\mathbf{x}_{i}, y_{i}\right)\right\}_{i=1}^{n}$ and sufficient unlabeled data set by $D_{U}=\left\{\mathbf{x}_{i}\right\}_{i=1}^{m}$, $n<<m$.

The new setting we study is more complex than the previous settings. The biggest challenge is severely limited ID labeled data. Separating ID and OOD data from unlabeled data is highly dependent on a prior model, which is trained by labeled data only. Due to limited ID labeled data, the prior model tends to be not effective enough in our setting. Therefore, separating ID and OOD data from unlabeled data is challenging, which causes poor performance of OOD detection. At the same time, this new setting is more realistic than the previous settings. In many real-world applications, collecting large-scale ID labeled data is costly and time-consuming. In order to solve the new problem effectively, we propose \underline{T}opological \underline{S}tructure \underline{L}earning~(TSL).

\subsection{Topological Space Representation Learning}\label{sec:Preliminary}
In the new setting, the number of labeled data is limited, which makes feature extraction more difficult. Representation learning by supervised learning is impracticable due to the limited number of labeled data. Large-scale unlabeled data is easy to collect. We use a self-supervised learning strategy to obtain reliable features based on all labeled and unlabeled data. Actually, contrastive learning can be considered as an extension of spectral clustering that can help extract features and learn the initial connections between instances \cite{haochen2021provable}. Concretely, we employ a simple contrastive learning method called SimCLR \cite{chen2020simple} on labeled data in $D_{L}$ and unlabeled data in $D_{U}$. SimCLR randomly samples a minibatch of $N$ examples and obtains $2N$ augmented examples by using two different augmentation strategies over the original $N$ examples. Then, SimCLR defines the contrastive prediction task as follows,
\begin{equation}
\mathcal{L}=\frac{1}{2 N} \sum_{k=1}^{N}[\ell(2 k-1,2 k)+\ell(2 k, 2 k-1)]\,,
\end{equation}
where $2k-1$ and $2k$ denote the two augmented examples from the same original example and are considered as a positive pair, similar to \cite{chen2017sampling}. SimCLR does not sample negative examples explicitly but treats the other $2(N-1)$ augmented examples within a minibatch as negative examples. $\ell$ denotes contrast loss, which is defined by
\begin{equation}
\ell_{i, j}=-\log \frac{\exp \left(\operatorname{sim}\left(\boldsymbol{z}_{i}, \boldsymbol{z}_{j}\right) / \tau\right)}{\sum_{k=1}^{2 N} \mathds{1}_{[k \neq i]} \exp \left(\operatorname{sim}\left(\boldsymbol{z}_{i}, \boldsymbol{z}_{k}\right) / \tau\right)}\,,
\end{equation}
where $z_{i}$ and $z_{j}$ are two different feature representations generated from an example $\mathbf{x}$ after the first three modules of SimCLR, $\operatorname{sim}\left(z_{i}, z_{j}\right)$ denotes cosine similarity of $z_{i}$ and $z_{j}$, and $\tau$ denotes a temperature parameter.

\subsection{Topological Connection Mining}\label{sec:Positive Negative Pair Mining}
To mine reliable connections of edges from limited ID labeled data and unlabeled data for the topological space and prepare for the next topological structure reconstruction, TSL uses all the reliable features extracted by SimCLR to mine the positive and negative pairs among data in $D_{L}\cup D_{U}$. Although the existing work \cite{zhou2021step} solves the problem of limited weakly-supervised OOD detection by positive and negative pairs mining, \cite{zhou2021step} suffers from noisy negative pairs and unreliable positive pairs. Our approach greatly alleviates the problems by reliable positive and negative pair mining.

The existing work, STEP \cite{zhou2021step} defines positive pairs as 
\begin{equation}
\mathcal{P}_{p}=\left\{\left(\mathbf{x}_{i}, \mathbf{x}_{j}\right) \mid \mathbf{x}_{i} \in \{D_{L}\cup D_{U}\}, \mathbf{x}_{j} \in \mathcal{B}_{K}\left(\mathbf{x}_{i}\right)\right\}\,,
\end{equation}
where $\mathbf{x}_{i}$ is sampled from $D_{L}\cup D_{U}$, and $\mathbf{x}_{j}$ is sampled from $\mathcal{B}_{K}\left(\mathbf{x}_{i}\right)$. $\mathcal{B}_{K}\left(\mathbf{x}_{i}\right)$ denotes the set of $K$ nearest neighbors of the example $\mathbf{x}_{i}$ which is defined by the KNN algorithm based on the Mahalanobis distance. Then for each instance, STEP considers all other instances except itself as its negative instances during the process of negative pairs mining by 
\begin{equation}
\mathcal{N}_{p}=\left\{\left(\mathbf{x}_{i}, \mathbf{x}_{j}\right) \mid i \neq j\right\}\,,
\end{equation}
where $\mathbf{x}_{i}$ and $\mathbf{x}_{j}$ are sampled from $D_{L}\cup D_{U}$.


\subsubsection{Positive Pair Mining} \label{ssec:Positive Negative Pair Mining}

\begin{figure}[t]
\setlength{\abovecaptionskip}{0.cm}
	\centering
	\includegraphics[width=0.7\textwidth]{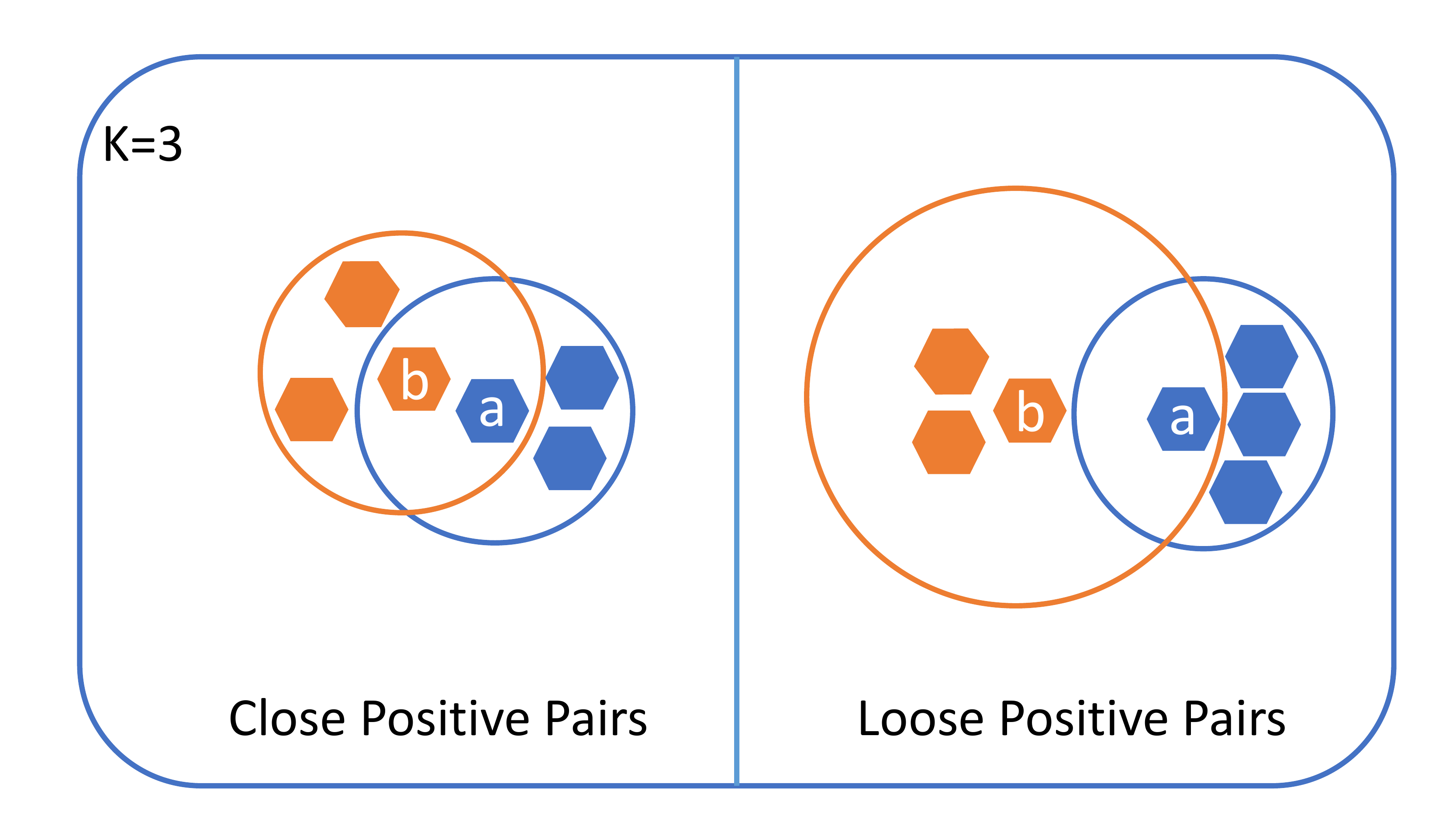}
	\caption{Mining of positive pairs. The left and right diagrams represent how TSL mines close positive pairs and loose positive pairs, respectively when $K$ is 3. In the left diagram, example `a' is one of the $K$ nearest neighbors of example `b', and example `b' is one of the $K$ nearest neighbors of example `a', so `a' and `b' are close positive pair. On the contrary, in the right diagram, example `a' is one of the $K$ nearest neighbors of example `b', but example `b' is not, so `a' and `b' is a loose positive pair.
	}
	\label{fig3}
\end{figure}

The mined positive pairs are one-sided and rough in STEP. An example gives the same degree of trust to all its $K$ nearest neighbors and considers all the $K$ nearest neighbors as its positive instances. Such a connection of positive pairs is not reliable. Fig.~\ref{fig3} demonstrates that the neighbor connection of `a' on the left is more intimate than that on the right. However, according to STEP, the connection of `a' and `b' is treated equally on both the left and right sides, which is problematic. Obviously, if two instances are each other's positive instances, then such a connection of positive pairs should be more reliable.

TSL notes that the positive pairs can be further subdivided into positive pairs with different credibility for better mining of positive pairs. Then TSL defines \emph{close positive pairs} and \emph{loose positive pairs} according to reciprocal and non-reciprocal neighbors. More specifically, we divide the $K$ nearest neighbors provided by the KNN algorithm according to Mahalanobis distance for each example into reciprocal neighbors and non-reciprocal neighbors. If two instances are both in each other's $K$ nearest neighbors, then we judge that the two instances are reciprocal neighbors with each other, which means that the two instances are close positive pairs, which is defined by
\begin{equation}
\mathcal{P}_{c}=\left\{\left(\mathbf{x}_{i}, \mathbf{x}_{j}\right) \mid \mathbf{x}_{i} \in \mathcal{B}_{K}\left(\mathbf{x}_{j}\right) \wedge \mathbf{x}_{j} \in \mathcal{B}_{K}\left(\mathbf{x}_{i}\right)\right\}\,,
\end{equation}
where $\mathbf{x}_{i}$ and $\mathbf{x}_{j}$ are randomly sampled from the feature space. The process of mining close positive pairs is clearly shown in the left diagram of Fig.~\ref{fig3}.
On the contrary, if example `a' is one of $K$ nearest neighbors of example `b', but example `b' is not one of $K$ nearest neighbors of example `a', we conclude that `a' and `b' are non-reciprocal neighbors with each other, which means the two instances are loose positive pairs, which is defined by

\begin{equation}
\mathcal{P}_{l}=\left\{\left(\mathbf{x}_{i}, \mathbf{x}_{j}\right) \mid \mathbf{x}_{i} \in \mathcal{B}_{K}\left(\mathbf{x}_{j}\right) \oplus \mathbf{x}_{j} \in \mathcal{B}_{K}\left(\mathbf{x}_{i}\right)\right\}\,,
\end{equation}
where $\mathbf{x}_{i}$ and $\mathbf{x}_{j}$ are randomly sampled from the feature space. The right diagram of Fig.~\ref{fig3} presents how TSL mines loose positive pairs. TSL subdivides positive pair connection by defining close positive pairs and loose positive pairs:
\begin{equation}
\mathcal{P}_{p}=\mathcal{P}_{l}+\mathcal{P}_{c}\,.
\end{equation}
\subsubsection{Negative Pair Mining} \label{ssec:Negative Negative Pair Mining}

The negative pairs screening has apparent drawbacks in STEP. For example, STEP considers two examples from the same class as a negative pair, which causes ineffective class distribution learning. To obtain reliable negative pairs, TSL defines a more targeted negative pairs instead of defining any two instances as negative pairs. More specifically, for each example $\mathbf{x}_{i}$, TSL progressively sorts all the remaining instances according to their Mahalanobis distance to $\mathbf{x}_{i}$. Then, TSL regards the instances ranked after the $\beta \times K$ as its negative instances. The negative example set of $\mathbf{x}_{i}$ can be expressed as
\begin{equation}
\mathcal{NB}_{\beta \times K }\left(\mathbf{x}_{i}\right)=\left\{\mathbf{x}_{j} \mid  \mathbf{x}_{j} \notin \mathcal{B}_{\beta \times K }\left(\mathbf{x}_{i}\right)\right\}\,,
\end{equation}
where $\mathcal{B}_{\beta \times K}\left(\mathbf{x}_{i}\right)$ denotes the $\beta \times K$ closest neighbors of $\mathbf{x}_{i}$. $\mathbf{x}_{i}$ and each of $\mathcal{NB}_{\beta \times K}\left(\mathbf{x}_{i}\right)$ form a negative pair. The set of negative pairs can be denoted by

\begin{figure}[t]
\setlength{\abovecaptionskip}{0.cm}
	\centering
	\includegraphics[width=0.7\textwidth]{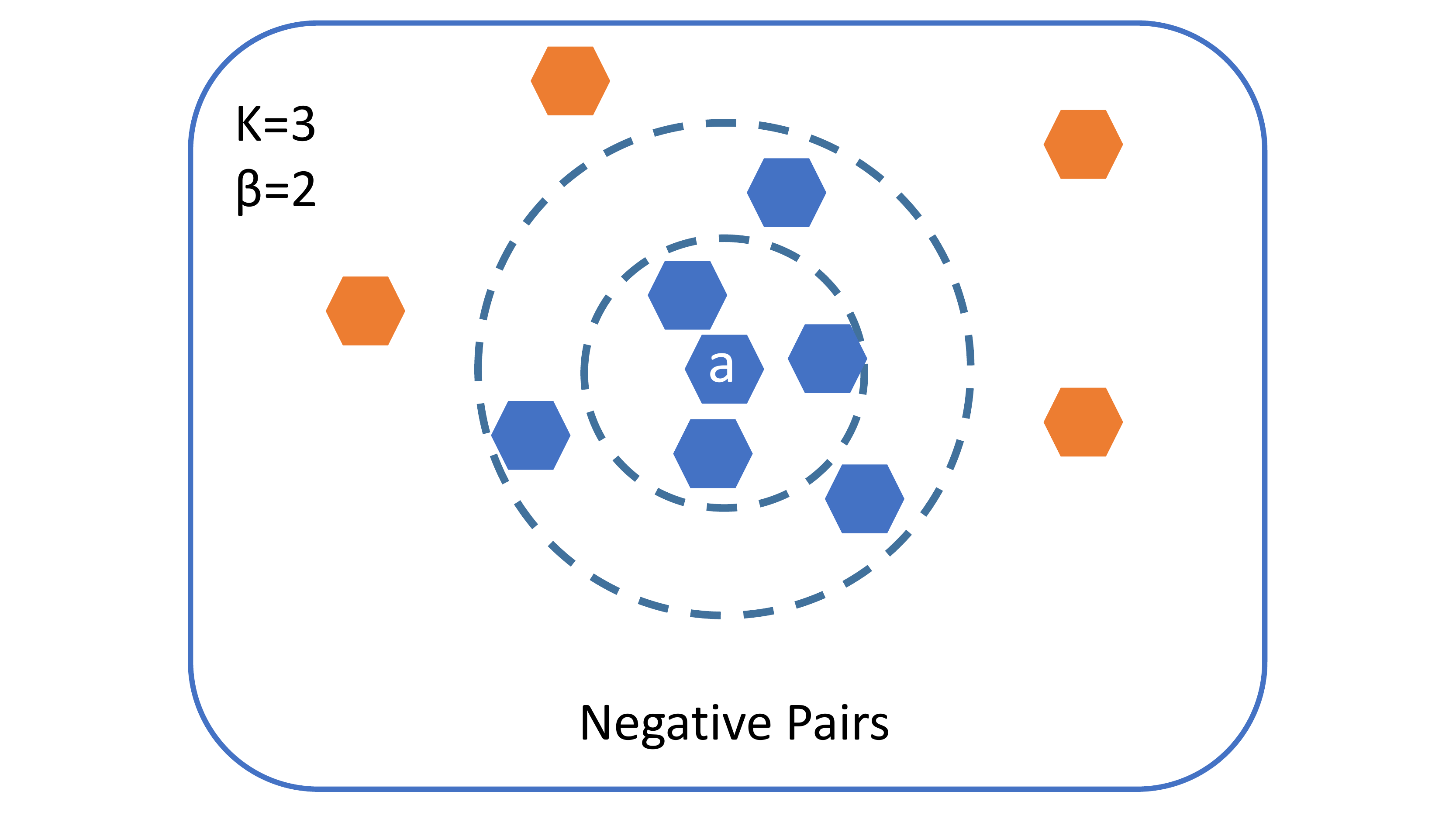}
	\caption{Mining of negative pairs. The figure defines $K$=3 and $\beta$=2, so all of the instances in the blue border are all $\beta \times K$ nearest neighbors of example `a' and the other instances in the yellow border are all negative instances of `a'.
	}
	\label{fig4}
\end{figure}

\begin{equation}
\mathcal{N}_{\beta p}=\left\{\left(\mathbf{x}_{i}, \mathbf{x}_{j}\right) \mid \mathbf{x}_{i} \in \{D_L \cup D_U\}, \mathbf{x}_{j} \in \mathcal{NB}_{\beta\times K}\left(\mathbf{x}_{i}\right)\right\}\,,
\end{equation}
where $\mathbf{x}_{i}$ is randomly sampled from $D_L \cup D_U$, and $\mathbf{x}_{j}$ is randomly sampled from $\mathcal{NB}_{\beta\times K}\left(\mathbf{x}_{i}\right)$. Fig.~\ref{fig4} illustrates how TSL mines negative pairs well. Such a negative pair mining method can alleviate the problem of defining positive pairs and instances from the same ID category as negative pairs to some extent.

\subsection{Topological Structure Reconstruction} \label{sec:Topological Space Reconstruction}
To increase the separability of ID and OOD instances, TSL reconstructs the topological structure in a new topological space by training the projector $\mathbf{P}$. The role of $\mathbf{P}$ is to project the features to a new feature space. When constructing the topological structure, TSL performs two steps. The first is the maintenance of the topological skeleton. The second is to extend and reconstruct the whole topological structure on the basis of the topological skeleton.

\subsubsection{Topological Skeleton Maintenance}\label{sec:Topological Skeleton Maintenance}

In the new topological space, TSL reduces the intra-class Mahalanobis distance of limited ID labeled data to maintain the topological skeleton. Therefore, in the new topological space, these instances form $k$ small clusters according to the ground truth labels, which further influences the distribution of all instances and makes the decision boundary between ID and OOD clearer. During the implementation, TSL considers any two instances of the same class from $D_{L}$ as a positive pair, performs metric learning of positive pairs by training the projector $\mathbf{P}$, and assigns a smaller coefficient than that of close positive pairs. TSL conducts $\mathcal{L}_{a}$ to aggregate ID intra-class labeled data. $\mathcal{L}_{a}$ is defined by 

\begin{equation}\label{la}
\mathcal{L}_{a}  =\sum_{\left(\mathbf{x}_{p}, \mathbf{x}_{q}\right)\in \mathcal{P}_{a}} \max \left(0,\left\|\mathbf{P} \mathbf{x}_{p}-\mathbf{P} \mathbf{x}_{q}\right\|_{2}-\lambda_1 \mathcal{M D}\left(\mathbf{x}_{p}, \mathbf{x}_{q}\right)\right)\ ,
\end{equation}
where $\lambda_1$ is a coefficient with the constraint of $\lambda_1 < \lambda_2$, and $\mathcal{P}_{a}$ is ID labeled positive pair set. $\mathcal{P}_{a}$ is defined by
\begin{equation}
\mathcal{P}_{a}=\left\{ (\mathbf{x}_{i}, \mathbf{x}_{j} ) \mid \mathbf{x}_{i}, \mathbf{x}_{j} \in D_L, y_i=y_j, \mathbf{x}_{i} \neq \mathbf{x}_{j} \right\}\,,
\end{equation}
where $\mathcal{P}_{a}$ denotes Intra-class positive pairs in $D_{L}$.

Compared to TSL, the existing method STEP does not fully use the valuable ID labeled data. STEP only uses the labeled instances when measuring known class centers and when measuring the covariance of the Mahalanobis distance, which causes a tremendous waste of the information carried by the ID labeled instances, especially considering the high cost of obtaining those labeled data.

\subsubsection{Topological Skeleton Extension}\label{sec:Topological Skeleton Extension} 

In the new topological space, based on the topological skeleton, TSL extends the topological structure with the positive and negative pairs mined before. Specifically, in the new feature space, TSL adjusts the distances of close positive and loose positive pairs while separating negative pairs as much as possible.

For positive pairs, TSL gives different confidence levels to close positive and loose positive pairs by imposing different coefficients when maintaining topological structure dynamically. Concretely, TSL imposes a lower coefficient on close positive pairs because they are more credible neighbors to each other. Close positive pairs ought to be closer to each other than loose positive pairs. Conversely, TSL imposes a higher coefficient on loose positive pairs because the connection of loose positive pairs is farther than close positive pairs. We construct $\mathcal{L}_{c}$ and $\mathcal{L}_{l}$ to implement the dynamic topological adjustment on the positive pairs. $\mathcal{L}_{c}$ is defined by
\begin{equation}\label{lc}
\mathcal{L}_{c}  =\sum_{(\mathbf{x}_{i}, \mathbf{x}_{n}) \in \mathcal{P}_{c}} \max \left(0,\left\|\mathbf{P} \mathbf{x}_{i}-\mathbf{P} \mathbf{x}_{n}\right\|_{2}-\lambda_2 \mathcal{M D}\left(\mathbf{x}_{i}, \mathbf{x}_{n}\right)\right)\\,
\end{equation}
where $\lambda_2$ is the coefficient for close positive pairs, $\mathbf{P} \mathbf{x}_{i}$ and $\mathbf{P} \mathbf{x}_{n}$ denote the projected features in the new feature space, and $\mathcal{M D}\left(\mathbf{x}_{i}, \mathbf{x}_{n}\right)$ denotes Mahalanobis distance between $\mathbf{x}_{i}$ and $\mathbf{x}_{n}$, which is defined by
\begin{equation}
\mathcal{M D}\left(\mathbf{x}_{i}, \mathbf{x}_{n}\right)=\sqrt{\left(\mathbf{x}_{i}-\mathbf{x}_{n}\right)^{\top} \hat{\boldsymbol{\Sigma}}^{-1}\left(\mathbf{x}_{i}-\mathbf{x}_{n}\right)}\,,
\end{equation}
where $\hat{\boldsymbol{\Sigma}}$ denotes covariance matrix of features of $D_{L}$. $\mathcal{L}_{l}$ is defined by
\begin{equation}\label{ll}
\mathcal{L}_{l}  =\sum_{(\mathbf{x}_{i}, \mathbf{x}_{n}) \in \mathcal{P}_{l}} \max \left(0,\left\|\mathbf{P} \mathbf{x}_{i}-\mathbf{P} \mathbf{x}_{n}\right\|_{2}-\lambda_3 \mathcal{M D}\left(\mathbf{x}_{i}, \mathbf{x}_{n}\right)\right)\\,
\end{equation}
where $\lambda_3$ is the coefficient for loose positive pairs, and $\lambda_2 < \lambda_3$. More details about $\lambda_2, \lambda_3$  will be introduced in the experiment setup.

For negative pairs, TSL constructs $\mathcal{L}_{f}$ as follows,
\begin{equation} \label{lf}
\mathcal{L}_{f}  =\sum_{{(\mathbf{x}_{i}, \mathbf{x}_{j}) \in \mathcal{N}_{\beta p}}}\max \left(0, M -\left\|\mathbf{P} \mathbf{x}_{i}-\mathbf{P} \mathbf{x}_{j}\right\|_{2}\right)\,,
\end{equation}
where $M$ denotes margin. The larger $M$ is, the farther the negative pairs are.

\subsection{Optimizing Objective}\label{sec:Optimizing Objective}
Considering Eqs.~\eqref{la}, \eqref{lc}, \eqref{ll} and \eqref{lf} , we state the overall loss $\mathcal{L}$ as
\begin{equation}
   \mathcal{L} = \mathcal{L}_a + \mathcal{L}_c + \mathcal{L}_l + \mathcal{L}_f\,.
\label{equoverall}
\end{equation}
Then we train the projector $\mathbf{P}$ for projecting the original topological space to the new topological space by minimizing the overall loss $\mathcal{L}$.

\subsection{Scoring Function Designing}\label{sec:Scoring Function Designing}
During inference, we use the output of projector $\mathbf{P}$ for OOD detection. In particular, given a test input $\mathbf{x}^*$, the OOD uncertainty score is given by
\begin{equation}
\operatorname{MS}(\mathbf{x}^*)= -\min _{c \in\left\{1, 2, \ldots, K\right\}} \left\|\mathbf{P} \mathbf{x}^*-\mathbf{P}\boldsymbol{\mu}_{c}\right\|_{2} \,,
\end{equation}
where $K$ denotes the number of ID classes, and $\boldsymbol{\mu}_{c}$ denotes $c$-th class centers of ID labeled data.

Setting a threshold $\gamma$ for $\operatorname{MS}(\cdot)$ can help the model distinguish between ID and OOD. With the role of score function $\operatorname{MS}(\cdot)$ and threshold $\gamma$, we can redefine the decision function $G$:
\begin{equation}
G_{\gamma}(\mathbf{x})= \begin{cases}1 & \operatorname{MS}(\mathbf{x}) \geq \gamma ,\\ 0 & \operatorname{MS}(\mathbf{x})<\gamma.\end{cases}
\end{equation}
The principle of choosing the threshold $\gamma$ is usually to allow the decision function $G$ to classify the vast majority of ID instances correctly (e.g., 95\%). The overall algorithm is summarized in Algorithm~\ref{Al}.

\begin{algorithm}[t] 
\label{Al}
	\caption{TSL (\underline{T}opological \underline{S}tructure \underline{L}earning)}
	\LinesNumbered 
	\KwIn{ID labeled data set $D_{L}$,
	unlabeled data set $D_{U}$,
	projector $\mathbf{P}$,
	number of neighbors $K$,
	coefficients $\lambda_1$, $\lambda_2$, $\lambda_3$, $\beta$, $M$,
	threshold $\gamma$
	}
	\KwOut{feature extractor $f$, projector $\mathbf{P}$}
	/*Topological Space Representation Learning:*/
	train $f$ via SimCLR on $\ D_{L}  \cup \ D_{U}$
	
	feed $\ D_{L}  \cup \ D_{U}$ into $f$ to get the reliable feature representations
	
	/*Topological Connection Mining:*/
	
	calculate $\hat{\Sigma}$ on reliable feature representations of $\ D_{L}$

	identify $\mathcal{B}_{k}\left(\mathbf{x}_{i}\right)$ for each example on reliable features via KNN with Mahalanobis distance
	
	mine $\mathcal{P}_{c}$, $\mathcal{P}_{l}$, $\mathcal{N}_{\beta p}$, and $\mathcal{P}_{a}$

	/*Topological Space Reconstruction:*/
	
	\For{$epoch\in\left\{1,2, \ldots\right.$, epoch $\left._{\max }\right\}$}{
		/*Topological Skeleton Maintenance:*/
		
	    compute $\mathcal{L}_a$ according to Eq.~\eqref{la} and $\mathcal{P}_{a}$\\
	    /*Topological Skeleton Extension:*/
	    
	    compute $\mathcal{L}_c$ according to Eq.~\eqref{lc} and $\mathcal{P}_{c}$\\
	    compute $\mathcal{L}_l$ according to Eq.~\eqref{ll} and $\mathcal{P}_{l}$\\
	    compute $\mathcal{L}_f$ according to Eq.~\eqref{lf} and $\mathcal{N}_{\beta p}$\\
	    
	    obtain overall loss $\mathcal{L}$ according to Eq.~\eqref{equoverall}\\
		optimize $\mathbf{P}$ via SGD according to $\mathcal{L}$\\

	}
	\Return $f$, $\mathbf{P}$
\end{algorithm}

\section{Experiments}\label{sec:experiment}
In this section, we validate the validity and robustness of our proposed method, TSL. Firstly, we introduce the experimental setup, including ID and OOD datasets, evaluation metrics, comparison methods, and implementation details. Secondly, to verify the validity and robustness of TSL, we compare it with the state-of-the-art under several representative datasets. Thirdly, we perform ablation experiments to verify the role of each module. Finally, we conduct numerous experiments to analyze the sensitivity of hyperparameters.
\label{sec:pagestyle}

\subsection{Experimental Setup}
\paragraph{In-distribution Dataset} Following \cite{zhou2021step}, we choose CIFAR-10 and CIFAR-100 \cite{krizhevsky2009learning} as the ID datasets. Both CIFAR-10 and CIFAR-100 include 50,000 training images and 10,000 testing images. The image size of the two data sets is 32 × 32. For CIFAR-10, there are ten classes in CIFAR-10. In the training phase, we randomly select 25 images from each class of CIFAR-10 to form $D_L$, and then place the rest of the images from the CIFAR-10's training set into $D_{U }^{in }$. For CIFAR-100, there are 100 classes. In the training phase, we randomly select four images from each class of CIFAR-100 to $D_L$ and then put the rest into $D_{U}$. In the testing phase, we randomly select 9,000 images in total from the testing set of CIFAR-10 or CIFAR-100 as $D_{test}^{in}$.

\paragraph{Out-of-distribution Dataset} Following \cite{zhou2021step}, we choose TinyImageNet-crop~(TINc), TinyImageNet-resize~(TINr),  LSUN-crop~(LSUNc) and LSUN-resize~(LSUNr) as OOD datasets. TINc and TINr are two variants of Tiny ImageNet~(TIN) \cite{deng2009imagenet}. TIN consists of 10,000 images from 200 classes, and it is a subset of ImageNet. TIN is changed into TINc or TINr by the operation of randomly cropping or downsampling each image to 32 × 32. LSUNc and LSUNr are two variants of the Large-scale Scene Understanding (LSUN) dataset \cite{yu2015lsun}. LSUN consists of 10,000 images from 10 classes. LSUN is changed into LSUNc or LSUNr by the same operation like TIN. We put the remaining data in OOD dataset into $D_{U}$ and $D_{test}^{out}$. Following STEP, we set $D_{U}^{out}$ and $D_{test}^{out}$ identical.

\paragraph{Evaluation Metrics} Our experiments use the following five test metrics for comparison: AUROC, FPR95, Detection Error, AUPR-In, and AUPR-Out. AUROC is the area obtained by integrating the ROC curve. The horizontal and vertical axes of the ROC curve represent FPR and TPR, respectively, and each point on the ROC curve reflects the value of FPR and TPR at different thresholds. The value of AUROC can be interpreted as the probability of ranking the positive example before the negative example, given a positive and a negative example at random. The higher the AUROC is, the better the performance of OOD detection. FPR95 denotes the probability of FPR when TPR equals 95$\%$. Detection Error indicates the minimum misclassification probability. We calculate Detection Error following \cite{zhou2021step}. AUPR denotes the area under the precision-recall curve, similarly to AUROC. AUPR-In and AUPR-Out regard ID and OOD data as positive instances, respectively. Among the above five metrics, we expect FPR95 and Detection Error to be as small as possible and AUROC, AUPR-In, and AUPR-Out to be as large as possible to obtain better OOD detection performance.

\paragraph{Comparison Methods} 
We compare our method with some classical and advanced OOD detection or PU learning methods to verify the effectiveness of TSL. \textbf{ADOA} \cite{zhang2018anomaly} is one of the two-stage PU learning methods. \textbf{ODIN} \cite{liang2017enhancing} is a baseline for OOD Detection, and it uses temperature scaling and input perturbation to amplify the ID/OOD separability. \textbf{MAH} \cite{lee2018simple} is characterized by measuring OOD confidence score of instances with Mahalanobis distance. \textbf{UOOD }\cite{yu2019unsupervised} is a representative of OOD detection with unlabeled Data, and \textbf{STEP} \cite{zhou2021step} is currently the best method that can be used to solve WSOOD.

\paragraph{Implementation Details} 
We choose Densenet-BC \cite{huang2017densely} as the backbone of SimCLR. Following \cite{zhou2021step}, we set $K$=12 and $M$=3. While maintaining the topological skeleton with ID labeled data, we set $\lambda_1$=0.1. While extending the topological skeleton, if the positive pair is a close positive pair, we set $\lambda_2$=0.5. On the contrary, if the positive pair is a loose positive pair, we set $\lambda_3$=6. And we set $\beta$=4000 for negative pairs. We repeat each set of experiments four times. Moreover, while training TSL, we set learning rate to 0.0003, epoch to 1500, and batch size to 128.

\begin{table}[t]
  \caption{Comparison of performance on several OOD benchmarks using five popular metrics. The results of ADOA, STEP, and TSL are from our experiments. We repeat experiments four times. The results of UOOD, ODIN, and MAH are from \cite{zhou2021step}.}  
  \centering
  \label{tab1}  
\scalebox{0.62}{
\begin{tabular}{c|ll|cccccc}
\toprule
\multicolumn{1}{l|}{Metrics}     & ID                        & OOD   & ADOA & ODIN          & MAH           & UOOD          & STEP                   & TSL(ours)                  \\ \midrule
\multirow{8}{*}{\rotatebox{90}{AUROC}}     \multirow{8}{*} {$\uparrow$}           & \multirow{4}{*}{\rotatebox{90}{CIFAR-10}}  & TINc  & 65.03 ± 4.12              & 81.00 ± 6.30  & 87.67 ± 2.47  & 90.46 ± 9.74  & 99.99 ± 0.00           & \textbf{100.00 ± 0.00} \\
                                 &                           & TINr  & 70.90 ± 2.09              & 59.10 ± 2.08  & 86.88 ± 0.87  & 84.67 ± 9.41  & 95.20 ± 0.47           & \textbf{98.51 ± 0.42}  \\
                                 &                           & LSUNc & 77.26 ± 3.06              & 76.17 ± 5.37  & 97.68 ± 0.09  & 96.92 ± 2.04  & \textbf{100.00 ± 0.00} & \textbf{100.00 ± 0.00} \\
                                 &                           & LSUNr & 74.28 ± 2.29              & 69.05 ± 3.49  & 90.41 ± 1.00  & 80.87 ± 24.45 & 99.34 ± 0.09           & \textbf{99.85 ± 0.03}  \\ \cline{2-9} 
                                 & \multirow{4}{*}{\rotatebox{90}{CIFAR-100}} & TINc  & 61.68 ± 4.62              & 61.65 ± 6.71  & 71.15 ± 2.20  & 98.34 ± 1.57  & \textbf{100.00 ± 0.00} & \textbf{100.00 ± 0.00} \\
                                 &                           & TINr  & 67.20 ± 2.48              & 54.46 ± 0.74  & 73.94 ± 1.79  & 84.80 ± 8.87  & 96.64 ± 0.62           & \textbf{98.93 ± 0.29}  \\
                                 &                           & LSUNc & 66.39 ± 2.45              & 46.99 ± 4.99  & 93.91 ± 3.41  & 97.49 ± 1.48  & 99.99 ± 0.00           & \textbf{100.00 ± 0.00} \\
                                 &                           & LSUNr & 58.89 ± 0.70              & 52.06 ± 2.24  & 78.45 ± 1.11  & 97.61 ± 0.55  & 97.87 ± 0.85           & \textbf{99.57 ± 0.06}  \\ \midrule
\multirow{8}{*}{\rotatebox{90}{FPR95}}     \multirow{8}{*} {$\downarrow$}  & \multirow{4}{*}{\rotatebox{90}{CIFAR-10}}  & TINc  & 94.50 ± 0.93              & 53.37 ± 10.55 & 44.17 ± 6.43  & 29.35 ± 30.05 & \textbf{0.00 ± 0.00}   & \textbf{0.00 ± 0.00}   \\
                                 &                           & TINr  & 85.70 ± 1.97              & 89.76 ± 1.45  & 58.57 ± 3.09  & 31.72 ± 11.50 & 20.70 ± 2.62           & \textbf{5.33 ± 1.43}   \\
                                 &                           & LSUNc & 81.06 ± 4.05              & 64.06 ± 9.12  & 7.73 ± 0.46   & 6.59 ± 3.22   & \textbf{0.00 ± 0.00}   & \textbf{0.00 ± 0.00}   \\
                                 &                           & LSUNr & 82.84 ± 2.31              & 76.89 ± 5.04  & 45.41 ± 3.87  & 32.69 ± 31.93 & 3.10 ± 0.45            & \textbf{0.52 ± 0.14}   \\ \cline{2-9} 
                                 & \multirow{4}{*}{\rotatebox{90}{CIFAR-100}} & TINc  & 97.44 ± 0.79              & 84.24 ± 8.02  & 90.15 ± 1.99  & 5.22 ± 5.59   & \textbf{0.00 ± 0.00}   & \textbf{0.00 ± 0.00}   \\
                                 &                           & TINr  & 94.70 ± 1.16              & 90.10 ± 0.46  & 80.55 ± 1.89  & 29.09 ± 15.68 & 13.54 ± 2.76           & \textbf{4.31 ± 1.16}   \\
                                 &                           & LSUNc & 91.06 ± 0.93              & 93.49 ± 2.42  & 24.93 ± 21.75 & 6.24 ± 3.80   & \textbf{0.00 ± 0.00}   & \textbf{0.00 ± 0.00}   \\
                                 &                           & LSUNr & 96.60 ± 0.49              & 89.79 ± 0.79  & 69.69 ± 2.42  & 4.92 ± 1.33   & 9.81 ± 4.48            & \textbf{1.76 ± 0.14}   \\ \midrule
\multirow{8}{*}{\rotatebox{90}{Detection Error}}     \multirow{8}{*} {$\downarrow$} & \multirow{4}{*}{\rotatebox{90}{CIFAR-10}}  & TINc  & 37.15 ± 3.16              & 25.53 ± 4.67  & 19.93 ± 2.63  & 11.59 ± 11.35 & \textbf{0.09 ± 0.01}   & 0.13 ± 0.03            \\
                                 &                           & TINr  & 34.18 ± 1.52              & 43.04 ± 1.48  & 20.14 ± 0.82  & 18.07 ± 5.55  & 11.26 ± 0.54           & \textbf{4.77 ± 0.82}   \\
                                 &                           & LSUNc & 29.14 ± 2.58              & 29.57 ± 3.82  & 6.28 ± 0.25   & 4.20 ± 2.12   & 0.11 ± 0.00            & \textbf{0.10 ± 0.03}    \\
                                 &                           & LSUNr & 31.92 ± 1.88              & 35.52 ± 2.46  & 16.23 ± 0.95  & 18.40 ± 15.68 & 3.76 ± 0.35            & \textbf{1.19 ± 0.20}   \\ \cline{2-9} 
                                 & \multirow{4}{*}{\rotatebox{90}{CIFAR-100}} & TINc  & 36.93 ± 3.42              & 40.95 ± 5.07  & 32.58 ± 1.64  & 3.67 ± 3.62   & \textbf{0.10 ± 0.04}   & 0.11 ± 0.04            \\
                                 &                           & TINr  & 35.10 ± 1.84              & 46.36 ± 0.56  & 31.09 ± 1.44  & 16.53 ± 7.87  & 8.85 ± 1.03            & \textbf{4.24 ± 0.70}   \\
                                 &                           & LSUNc & 37.26 ± 2.04              & 48.47 ± 1.61  & 11.20 ± 3.73  & 4.24 ± 2.34   & 0.24 ± 0.07            & \textbf{0.11 ± 0.02}   \\
                                 &                           & LSUNr & 41.54 ± 0.42              & 46.73 ± 0.66  & 27.33 ± 1.03  & 3.11 ± 0.78   & 7.10 ± 1.92            & \textbf{2.44 ± 0.16}   \\ \midrule
\multirow{8}{*}{\rotatebox{90}{AUPR-In}}     \multirow{8}{*} {$\uparrow$}         & \multirow{4}{*}{\rotatebox{90}{CIFAR-10}}  & TINc  & 70.58 ± 4.19              & 76.80 ± 8.20  & 85.35 ± 2.86  & 89.31 ± 10.05 & \textbf{99.99 ± 0.00}  & \textbf{99.99 ± 0.00}  \\
                                 &                           & TINr  & 72.19 ± 1.94              & 57.10 ± 2.11  & 86.79 ± 1.17  & 79.02 ± 12.17 & 94.58 ± 0.41           & \textbf{97.97 ± 0.60}  \\
                                 &                           & LSUNc & 79.74 ± 2.96              & 72.16 ± 6.60  & 96.70 ± 0.21  & 94.78 ± 4.07  & \textbf{100.00 ± 0.00} & \textbf{100.00 ± 0.00} \\
                                 &                           & LSUNr & 76.04 ± 2.46              & 65.37 ± 3.39  & 89.93 ± 1.23  & 79.41 ± 19.89 & 99.26 ± 0.08           & \textbf{99.81 ± 0.04}  \\ \cline{2-9} 
                                 & \multirow{4}{*}{\rotatebox{90}{CIFAR-100}} & TINc  & 69.65 ± 4.52              & 58.29 ± 5.01  & 71.18 ± 2.69  & 97.55 ± 2.04  & \textbf{100.00 ± 0.00} & \textbf{100.00 ± 0.00} \\
                                 &                           & TINr  & 71.52 ± 2.58              & 52.96 ± 0.59  & 70.95 ± 2.20  & 77.32 ± 9.81  & 96.03 ± 0.65           & \textbf{98.66 ± 0.39}  \\
                                 &                           & LSUNc & 68.31 ± 3.02              & 47.41 ± 2.86  & 92.26 ± 2.17  & 95.45 ± 2.32  & 99.99 ± 0.00           & \textbf{100.00 ± 0.00} \\
                                 &                           & LSUNr & 62.87 ± 0.73              & 50.47 ± 1.75  & 74.22 ± 1.14  & 95.53 ± 0.95  & 97.70 ± 0.86           & \textbf{99.46 ± 0.11}  \\ \midrule
\multirow{8}{*}{\rotatebox{90}{AUPR-Out}}     \multirow{8}{*} {$\uparrow$}        & \multirow{4}{*}{\rotatebox{90}{CIFAR-10}}  & TINc  & 58.47 ± 3.30              & 83.63 ± 5.11  & 88.67 ± 2.28  & 91.34 ± 8.69  & \textbf{99.99 ± 0.00}  & \textbf{99.99 ± 0.00}  \\
                                 &                           & TINr  & 67.40 ± 2.22              & 58.83 ± 1.77  & 84.26 ± 0.95  & 89.21 ± 6.22  & 95.63 ± 0.58           & \textbf{98.86 ± 0.31}  \\
                                 &                           & LSUNc & 73.46 ± 3.44              & 78.43 ± 5.12  & 98.16 ± 0.12  & 98.01 ± 1.18  & \textbf{100.00 ± 0.00} & \textbf{100.00 ± 0.00} \\
                                 &                           & LSUNr & 70.94 ± 2.33              & 70.51 ± 3.97  & 88.84 ± 1.20  & 84.45 ± 21.48 & 99.42 ± 0.08           & \textbf{99.88 ± 0.02}  \\ \cline{2-9} 
                                 & \multirow{4}{*}{\rotatebox{90}{CIFAR-100}} & TINc  & 53.68 ± 3.36              & 62.88 ± 7.90  & 65.14 ± 2.21  & 98.77 ± 1.23  & \textbf{100.00 ± 0.00} & \textbf{100.00 ± 0.00} \\
                                 &                           & TINr  & 59.91 ± 2.29              & 55.94 ± 0.71  & 71.57 ± 1.71  & 89.44 ± 6.96  & 97.05 ± 0.60           & \textbf{99.14 ± 0.23}  \\
                                 &                           & LSUNc & 61.93 ± 1.94              & 49.91 ± 4.42  & 93.77 ± 5.30  & 98.33 ± 0.99  & 99.99 ± 0.00           & \textbf{100.00 ± 0.00}  \\
                                 &                           & LSUNr & 53.41 ± 0.80              & 55.18 ± 1.56  & 78.19 ± 1.33  & 98.49 ± 0.37  & 98.05 ± 0.82           & \textbf{99.65 ± 0.04}  \\ \bottomrule
\end{tabular}}
\end{table}


\subsection{OOD Detection Performance}

Table~\ref{tab1} shows the OOD detection performance of TSL and baseline methods. The results show that PU learning methods, like ADOA, post-hoc-based methods, like ODIN and MAH, and fine-tuning-based methods with unlabeled data, like UOOD, fail to solve weakly-supervised OOD detection. PU learning and our problem have different assumptions about the proportion of positive instances in unlabeled data, so the performance of ADOA is not good. Post-hoc-based methods rely entirely on ID labeled data, so their performance is bound to drop when facing a severe lack of ID labeled data. For UOOD, due to the severe scarcity of ID labeled data, the empirical risk of treating all unlabeled instances as OOD data during training is no longer affordable. Therefore, the OOD detection performance of UOOD drops dramatically.

\begin{figure}[t]

\centering

\subfigure[]{
\begin{minipage}[t]{0.2\linewidth}
\centering
\includegraphics[height=3cm,width=2.8cm]{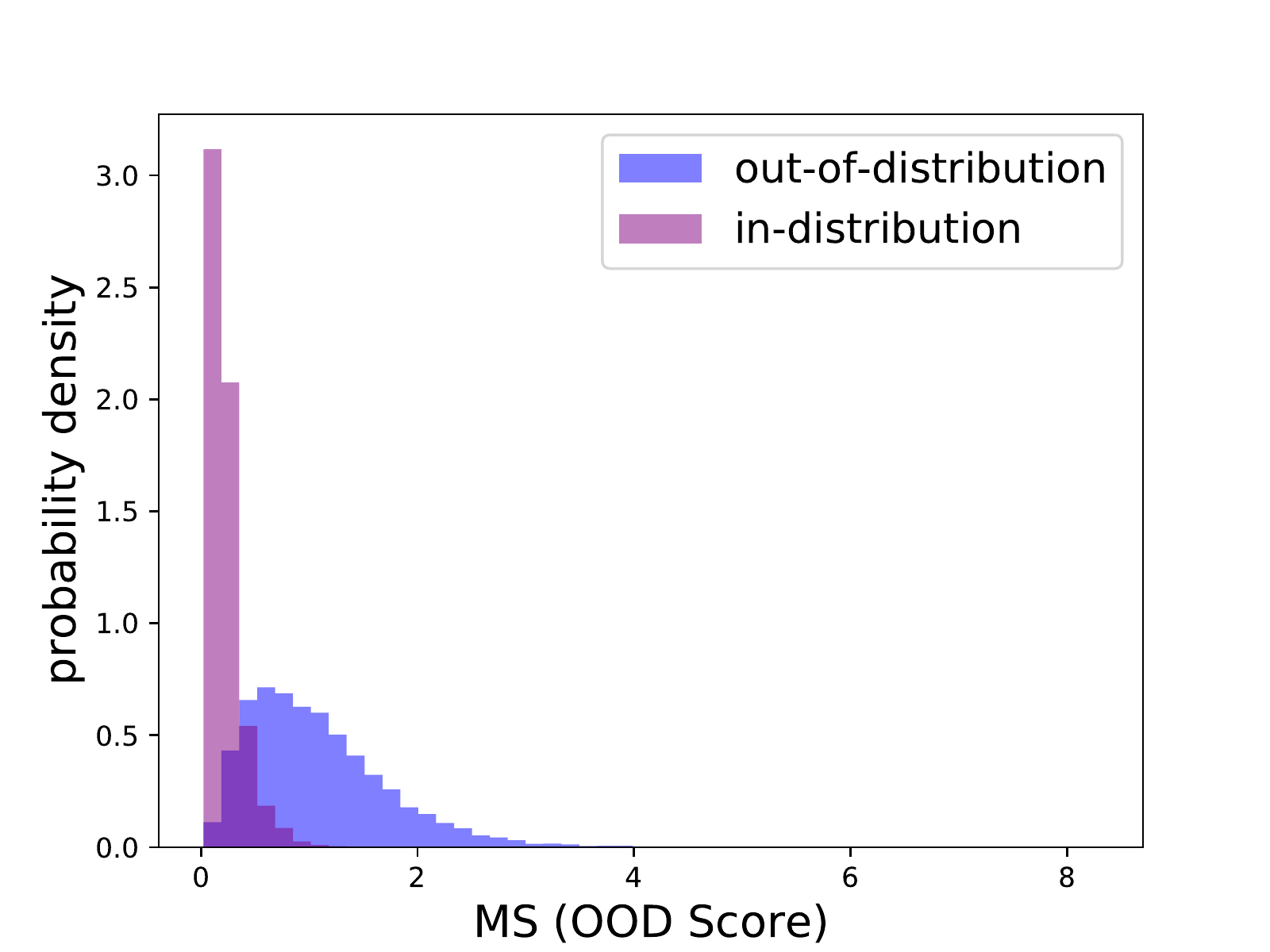}
\end{minipage}%
}%
\subfigure[]{
\begin{minipage}[t]{0.2\linewidth}
\centering
\includegraphics[height=3cm,width=2.8cm]{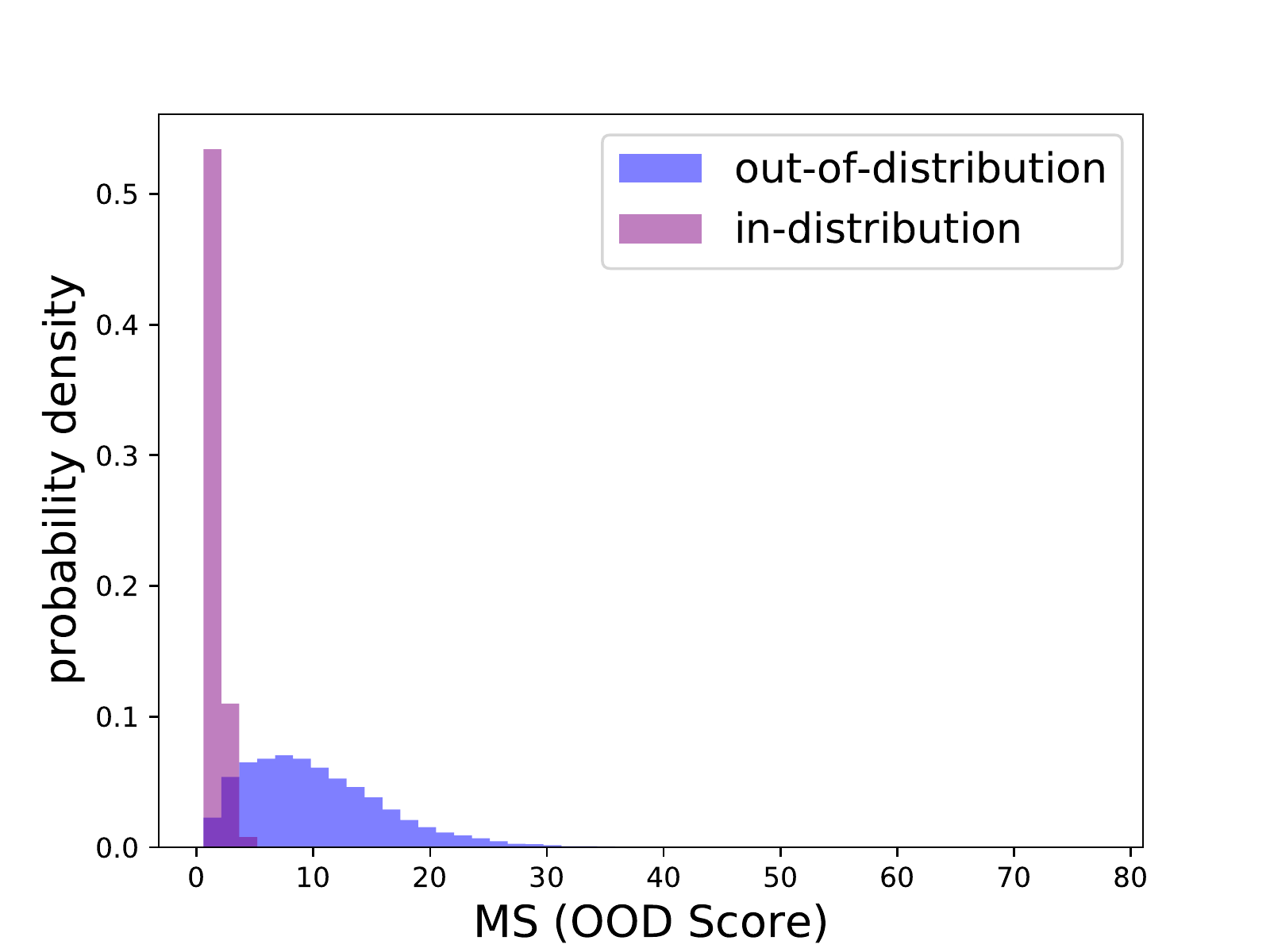}
\end{minipage}%
}%
\subfigure[]{
\begin{minipage}[t]{0.2\linewidth}
\centering
\includegraphics[height=3cm,width=2.8cm]{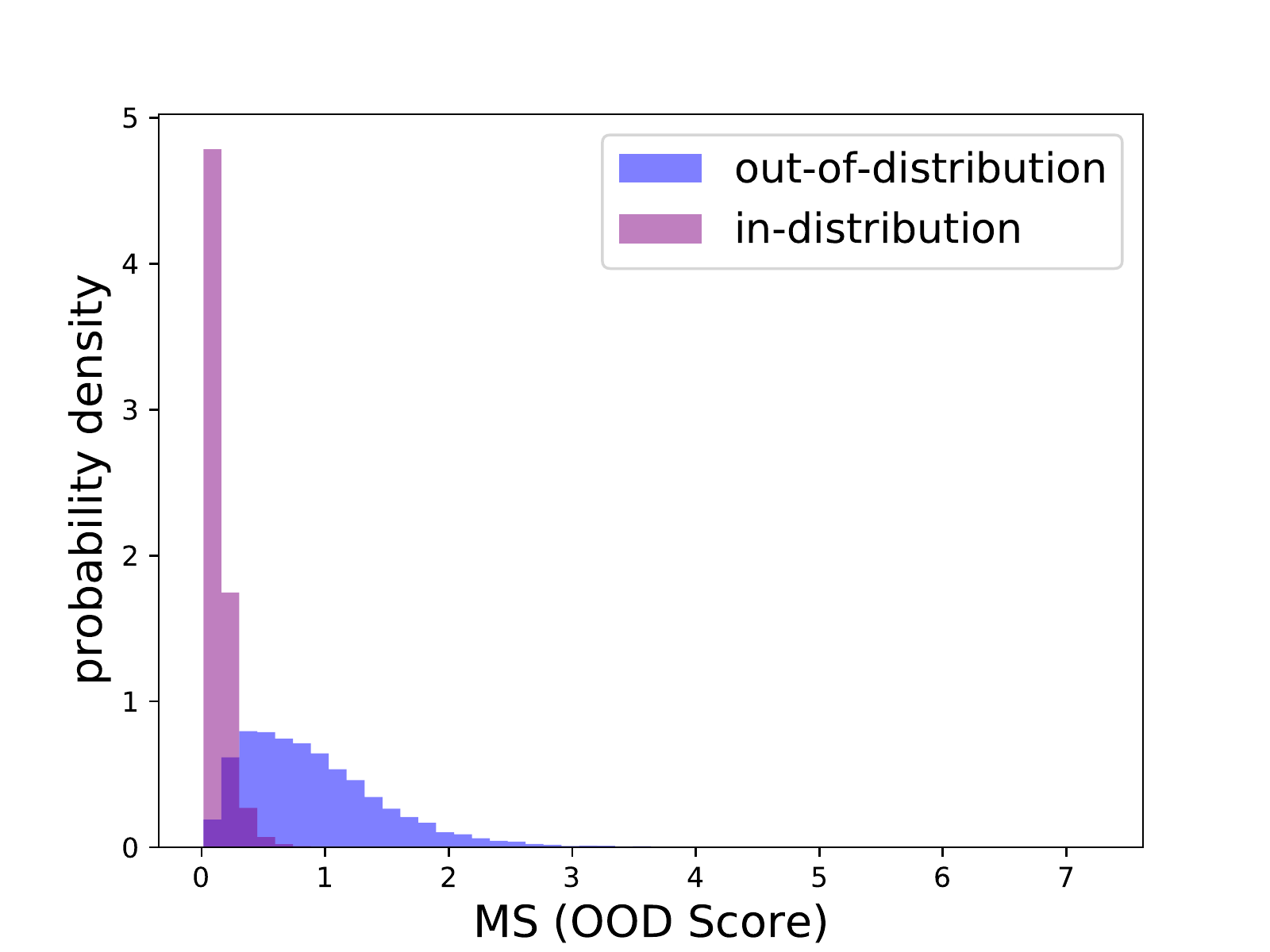}
\end{minipage}
}%
\subfigure[]{
\begin{minipage}[t]{0.2\linewidth}
\centering
\includegraphics[height=3cm,width=2.8cm]{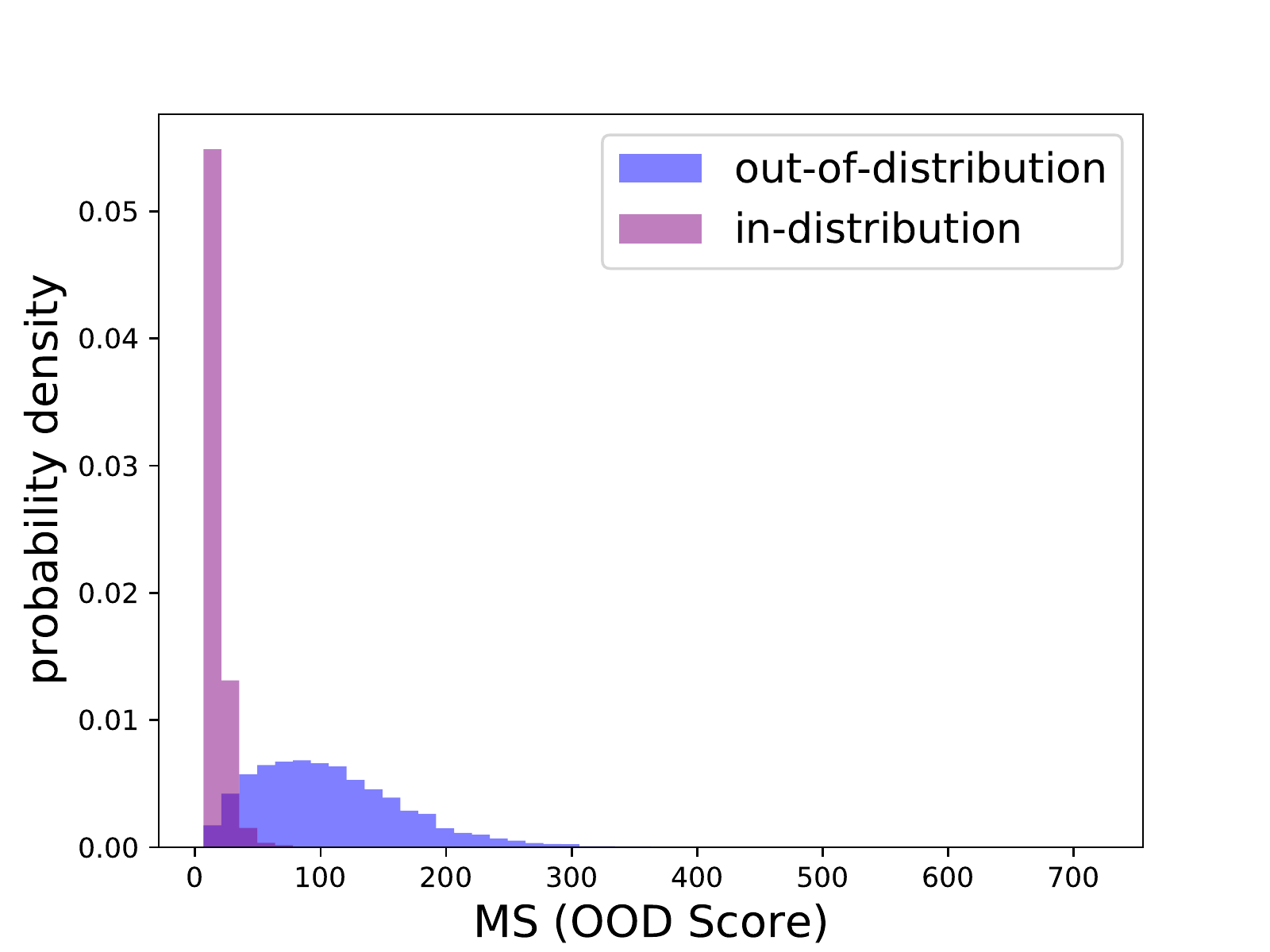}
\end{minipage}
}%
\centering
\caption{Comparison of the distribution of MS score for TSL and STEP on different benchmarks. (a): STEP on CIFAR-10 \& TINr. (b): TSL on CIFAR-10 \& TINr. (c): STEP on CIFAR-100 \& TINr. (d): TSL on CIFAR-100 \& TINr}
 \label{fig7} 
\end{figure}

STEP is the state-of-the-art OOD detection method when ID labeled data is limited. Surprisingly, TSL performs better than SETP on almost all datasets. Especially on difficult tasks, such as CIFAR-10 \& TINr, CIFAR-100 \& TINr, CIFAR-10 \& LSUNr, and CIFAR-100 \& LSUNr, TSL outperforms STEP by a large margin. We use the symbol \& to connect the ID dataset and the OOD dataset in a task. For example, TSL reduces the FPR95 and Detection Error by \textbf{15.37\%} and \textbf{6.49\%} compared to STEP on CIFAR-10 \& TINr, respectively. Through Fig.~\ref{fig7}, we can also observe that TSL separates ID and OOD instances more effectively compared to STEP. The above results verify the effectiveness of TSL. The composition of data set is shown in Table~\ref{composition of dataset}.

\begin{table}[t]
\caption{The composition of each data set in the experiment of OOD detection performance.}
\centering
\scalebox{0.65}{
\label{composition of dataset}
\scalebox{0.9}{\setlength{\tabcolsep}{2mm}{\begin{tabular}{ll|ccc}
\toprule
    & Dataset Combination   & $D_{L }$  & $D_{U }$    & $D_{test}$   \\ \midrule
ID  & CIFAR-10              & 250 & 49750 & 9000 \\
OOD & TINc/TINr/LSUNc/LSUNr & 0   & 9000  & 9000 \\ \midrule
ID  & CIFAR-100             & 400 & 49600 & 9000 \\
OOD & TINc/TINr/LSUNc/LSUNr & 0   & 9000  & 9000 \\ \bottomrule
\end{tabular}}}}
\end{table}

\subsection{Generalization of TSL}

\subsubsection{Generalizing to Unseen OOD Data}

\begin{table}[t]
\renewcommand{\arraystretch}{1.4}
  \caption{Performance comparison of STEP and TSL when the OOD data in the unlabeled dataset for training and the OOD dataset for testing do not come from the same distribution.}  
  \centering
  \label{tab2}  
\scalebox{0.65}{\setlength{\tabcolsep}{0.9mm}{\begin{tabular}{c|ccc|cc}
\toprule
Metrics                          & $D_{L }$\&$D_{U }^{in }$ \&$D_{test }^{in }$                        & $D_{U }^{out }$ & $D_{test }^{out }$ & STEP          & TSL                  \\ \midrule
\multirow{4}{*}{\rotatebox{90}{AUROC}}     \multirow{4}{*} {$\uparrow$}         & \multirow{2}{*}{CIFAR-10}  & TINc       & TINr      & 80.84 ± 2.12  & \textbf{87.32 ± 3.75}  \\
                                 &                           & TINr       & TINc      & 91.11 ± 0.79  & \textbf{93.78 ± 1.78}  \\
                                  \cline{2-6} 
                                 & \multirow{3}{*}{CIFAR-100} & TINc       & TINr      & 74.59 ± 3.18  & \textbf{82.29 ± 3.45}  \\
                                 &                           & TINr       & TINc      & 91.14 ± 2.51  & \textbf{94.62 ± 1.53}  \\\midrule
\multirow{4}{*}{\rotatebox{90}{FPR95}}     \multirow{4}{*} {$\downarrow$}  & \multirow{2}{*}{CIFAR-10}  & TINc       & TINr      & 58.66 ± 3.04  & \textbf{42.69 ± 11.15} \\
                                 &                           & TINr       & TINc      & 57.32 ± 2.78  & \textbf{34.30 ± 7.24}  \\
                     \cline{2-6} 
                                 & \multirow{3}{*}{CIFAR-100} & TINc       & TINr      & 72.31 ± 6.14  & \textbf{57.52 ± 7.45}  \\
                                 &                           & TINr       & TINc      & 48.68 ± 12.74 & \textbf{29.48 ± 8.50}  \\
                              \midrule
\multirow{4}{*}{\rotatebox{90}{Dection Error}}     \multirow{4}{*} {$\downarrow$} & \multirow{2}{*}{CIFAR-10}  & TINc       & TINr      & 25.08 ± 1.86  & \textbf{18.76 ± 3.80}  \\
                                 &                           & TINr       & TINc      & 14.85 ± 1.09  & \textbf{13.23 ± 2.78}  \\
                                \cline{2-6} 
                                 & \multirow{3}{*}{CIFAR-100} & TINc       & TINr      & 30.95 ± 3.03  & \textbf{24.16 ± 3.59}  \\
                                 &                           & TINr       & TINc      & 15.58 ± 3.01  & \textbf{11.95 ± 2.09}  \\
                             \midrule
\multirow{4}{*}{\rotatebox{90}{AUPR-In}}     \multirow{4}{*} {$\uparrow$}         & \multirow{2}{*}{CIFAR-10}  & TINc       & TINr      & 76.25 ± 2.42  & \textbf{84.67 ± 4.55}  \\
                                 &                           & TINr       & TINc      & 93.11 ± 0.72  & \textbf{94.54 ± 1.73}  \\
                              \cline{2-6} 
                                 & \multirow{3}{*}{CIFAR-100} & TINc       & TINr      & 71.53 ± 2.74  & \textbf{80.04 ± 3.85}  \\
                                 &                           & TINr       & TINc      & 92.56 ± 2.16  & \textbf{95.29 ± 1.37}  \\
                               \midrule
\multirow{4}{*}{\rotatebox{90}{AUPR-Out}}     \multirow{4}{*} {$\uparrow$}        & \multirow{2}{*}{CIFAR-10}  & TINc       & TINr      & 81.50 ± 1.80  & \textbf{86.39 ± 3.62}  \\
                                 &                           & TINr       & TINc      & 86.87 ± 1.01  & \textbf{92.80 ± 1.62}  \\
                               \cline{2-6} 
                                 & \multirow{3}{*}{CIFAR-100} & TINc       & TINr      & 73.74 ± 3.45  & \textbf{81.24 ± 3.04}  \\
                                 &                           & TINr       & TINc      & 87.59 ± 3.17  & \textbf{93.43 ± 1.74}          \\ \bottomrule
\end{tabular}}}

\end{table}

\begin{table*}[b]
  \caption{Performance comparison of STEP and TSL after dividing the OOD instances into $D_{U}^{out}$ and $D_{test}^{out}$ at a ratio of 8:2.}  
  \label{tab6} 
  \centering
\scalebox{0.73}{\setlength{\tabcolsep}{0.9mm}{\begin{tabular}{cc|c|ccccc}
\toprule
ID                         & OOD                   & Method & AUROC $\uparrow$                 & FPR95 $\downarrow$                & Detection Error $\downarrow$      & AUPR-In $\uparrow$               & AUPR-Out $\uparrow$              \\ \midrule
\multirow{2}{*}{CIFAR-10}  & \multirow{2}{*}{TINr} & STEP   & 93.44 ± 0.93          & 29.29 ± 3.63         & 13.56 ± 0.95         & 98.38 ± 0.24          & 80.12 ± 3.15          \\
                           &                       & TSL  & \textbf{98.65 ± 0.25} & \textbf{5.14 ± 0.94} & \textbf{4.72 ± 0.64} & \textbf{99.64 ± 0.08} & \textbf{96.70 ± 0.53} \\ \midrule
\multirow{2}{*}{CIFAR-100} & \multirow{2}{*}{TINr} & STEP   & 95.70 ± 0.92          & 16.95 ± 2.93         & 10.40 ± 1.39         & 98.88 ± 0.24          & 89.13 ± 2.12          \\
                           &                       & TSL  & \textbf{97.91 ± 0.51} & \textbf{7.62 ± 1.85} & \textbf{6.06 ± 1.03} & \textbf{99.45 ± 0.12} & \textbf{95.05 ± 1.35} \\ \bottomrule
\end{tabular}}}
\end{table*}

In the previous experimental setup, $D_{U}^{out}$ and $D_{test}^{out}$ are totally identical, which follows the setting of STEP. However, such an assumption is difficult to meet in the real world. For example, in autonomous driving, we cannot provide the model with all the OOD instances encountered in the future during the training phase. To explore the generalization ability of TSL when detecting unseen OOD data, we conduct two expanded experiments. In the first experiment, we randomly divide OOD data into $D_{U}^{out}$ and $D_{test}^{out}$ according to the ratio of 8:2. Following this experimental setup, we conduct experiments on CIFAR-10 \& TINr and CIFAR-100 \& TINr. By Table~\ref{tab6}, we observe that compared to STEP, TSL reduces the average FPR95 by \textbf{24.15\%} and \textbf{9.33\%} on CIFAR-10 \& TINr and CIFAR-100 \& TINr, respectively. The experiment results show that TSL is far ahead of STEP in terms of OOD generalization ability on unseen OOD data of the same classes. In the second experiment, $D_{U}^{out}$ is no longer the same as $D_{test}^{out}$. For example, we consider TINc as $D_{U}^{out}$ and TINr as $D_{test}^{out}$. From the results in Table~\ref{tab2}, we can observe that OOD generalization capability of TSL is significantly better than STEP. And we also observe that the task difficulty of generalizing TINr to TINc is lower than generalizing TINc to TINr. Whether the generalization task is easy or difficult, our proposed TSL outperforms STEP comprehensively, which verifies the generalization ability of TSL.


\begin{table}[h]
\caption{The performance of the near OOD detection experiment when we take CIFAR-10's testing set and all unlabeled data during training as the test set respectively.}  
\centering
\label{tab34}
\scalebox{0.65}{
\begin{tabular}{lcc|cc}
\toprule
        & \multicolumn{2}{c|}{$D_{U} \neq D_{test}$} & \multicolumn{2}{c}{$D_{U} = D_{test}$} \\ \midrule
Metrics         & STEP           & TSL                     & STEP                & TSL                          \\ \midrule
AUROC $\uparrow$           & 90.73 ± 0.35   & \textbf{93.91 ± 0.48}   & 90.96 ± 0.51        & \textbf{94.49 ± 1.34}        \\
FPR95 $\downarrow$           & 41.91 ± 2.33   & \textbf{24.90 ± 0.86}   & 39.44 ± 3.52        & \textbf{24.00 ± 5.81}        \\
Detection Error $\downarrow$ & 16.00 ± 0.23   & \textbf{11.69 ± 0.52}   & 15.75 ± 0.27        & \textbf{11.67 ± 1.62}        \\
AUPR-In $\uparrow$         & 93.13 ± 0.44   & \textbf{95.34 ± 0.44}   & 92.86 ± 0.32        & \textbf{95.65 ± 1.04}        \\
AUPR-Out $\uparrow$        & 84.72 ± 0.50   & \textbf{89.69 ± 1.25}   & 86.01 ± 1.71        & \textbf{92.34 ± 1.97}        \\  \bottomrule
\end{tabular}}
\end{table}

\subsubsection{Near OOD Detection}

To further explore the generalizability of TSL, we conduct the following experiments on CIFAR-10. CIFAR-10 contains ten classes in total. We consider six classes of animals as ID classes and four classes of transportation as OOD classes. We use the training set of CIFAR-10 to construct $D_{L}$ and $D_{U}$. We randomly sample 25 instances from each of the six animal classes' training sets as $D_{L}$ and treat the remaining 4975 instances of each of the six classes as $D_{U}^{in}$. Then we randomly sample 4975 instances from each of the four animal classes' training sets respectively as $D_{U}^{out}$.

We first build $D_{test}$ mixed with ID and OOD data using the test set of CIFAR-10, which defines the test set of six classes of animals as $D_{test}^{in}$ and defines the test set of four classes of transportation as $D_{test}^{out}$. In the left half of Table~\ref{tab34}, we show that TSL reduces the average FPR95 by \textbf{17.01\%} compared to STEP. Then we recover all the labels in $D_{U}$ and directly use $D_{U}$ as $D_{test}$. In the right half of Table~\ref{tab34}, we observe that TSL reduces the average FPR95 by \textbf{15.44\%} compared to STEP. As seen from the above two experiments, in the near OOD setting, the performance of TSL is significantly improved compared with STEP, regardless of whether or not the test set is visible in the training phase. At the same time, the performance of TSL is not significantly degraded when the test set is not visible while training.

\begin{table*}[H]
  \caption{Performance comparison of STEP and TSL after dividing the OOD instances into $D_{U}^{out}$ and $D_{test}^{out}$ at a ratio of 8:2.}  
  \label{tab6} 
\begin{tabular}{cc|c|ccccc}
\toprule
ID                         & OOD                   & Method & AUROC $\uparrow$                 & FPR95 $\downarrow$                & Detection Error $\downarrow$      & AUPR-In $\uparrow$               & AUPR-Out $\uparrow$              \\ \midrule
\multirow{2}{*}{CIFAR-10}  & \multirow{2}{*}{TINr} & STEP   & 93.44 ± 0.93          & 29.29 ± 3.63         & 13.56 ± 0.95         & 98.38 ± 0.24          & 80.12 ± 3.15          \\
                           &                       & TSL  & \textbf{98.65 ± 0.25} & \textbf{5.14 ± 0.94} & \textbf{4.72 ± 0.64} & \textbf{99.64 ± 0.08} & \textbf{96.70 ± 0.53} \\ \midrule
\multirow{2}{*}{CIFAR-100} & \multirow{2}{*}{TINr} & STEP   & 95.70 ± 0.92          & 16.95 ± 2.93         & 10.40 ± 1.39         & 98.88 ± 0.24          & 89.13 ± 2.12          \\
                           &                       & TSL  & \textbf{97.91 ± 0.51} & \textbf{7.62 ± 1.85} & \textbf{6.06 ± 1.03} & \textbf{99.45 ± 0.12} & \textbf{95.05 ± 1.35} \\ \bottomrule
\end{tabular}
\end{table*}

\subsection{Ablation Study}

We perform ablation experiments to verify the effectiveness of each of the three critical modules of TSL. As introduced in Section~\ref{sec:method}, TSL has three effective modules to improve the performance of OOD detection compared with STEP: positive pairs mining~(PPM), negative pairs mining~(NPM), and the topological skeleton
maintenance~(TSM) module. We perform ablation experiments on these three modules on CIFAR-10 \& TINr, CIFAR-10 \& LSUNr, CIFAR-100 \& TINr, and CIFAR-100 \& LSUNr, respectively, to show their contribution to TSL's performance. We add the three modules to STEP in turn to verify the effectiveness of each module. At the same time, to demonstrate the effectiveness of each module after stacking, we fix the random seed during the ablation experiment. From Table~\ref{tab:aba}, we observe that the model's performance on each combination of datasets constantly improves when adding PPM, NPM, and TSM in turn.

\begin{table}[t]
\setlength{\abovecaptionskip}{0.cm}
\caption{TSL's module ablation studies on different combinations of ID dataset and OOD dataset. In the table, PPM denotes the module of positive pairs mining, NPM denotes the module of negative pairs mining, and TSM denotes the module of topological skeleton
maintenance with limited ID labeled data.}
\centering
\label{tab5} 
\scalebox{0.65}{
\begin{tabular}{cc|ccc|ccccc}
\toprule
ID & OOD & PPM & NPM & TSM & AUROC $\uparrow$ & FPR95 $\downarrow$ & Detection Error $\downarrow$ & AUPR-In $\uparrow$ & AUPR-Out $\uparrow$\\\midrule
\multirow{4}{*}{CIFAR-10}& \multirow{4}{*}{TINr} &- & - & - & 95.36 & 20.41 & 11.16 & 94.75 & 95.85\\
&&$\checkmark$ & - & - & 96.87 & 13.38 & 8.72 & 96.31 & 97.31 \\
&&$\checkmark$ & $\checkmark$ & - & 98.06 & 7.36 & 6.11 & 97.30 & 98.48 \\
&&$\checkmark$ & $\checkmark$ & $\checkmark$ & \textbf{99.14} & \textbf{2.84} & \textbf{2.98} & \textbf{98.72} & \textbf{99.37}\\
\midrule
\multirow{4}{*}{CIFAR-10}& \multirow{4}{*}{LSUNr} &- & - & - & 99.12 & 4.34 & 4.64 & 99.04 & 99.21\\
&&$\checkmark$ & - & - & 99.71 & 1.21 & 1.97 & 99.67 & 99.76 \\
&&$\checkmark$ & $\checkmark$ & - & 99.78 & 0.98 & 1.68 & 99.74 & 99.81 \\
&&$\checkmark$ & $\checkmark$ & $\checkmark$ & \textbf{99.92} & \textbf{0.21} & \textbf{0.71} & \textbf{99.89} & \textbf{99.94}\\
\midrule
\multirow{4}{*}{CIFAR-100}& \multirow{4}{*}{TINr} &- & - & - & 96.96 & 12.24 & 8.27 & 96.36 & 97.41\\
&&$\checkmark$ & - & - & 98.00 & 7.99 & 6.42 & 97.50 & 98.34 \\
&&$\checkmark$ & $\checkmark$ & - & 98.16 &6.06 & 5.46 & 97.34 & 98.56 \\
&&$\checkmark$ & $\checkmark$ & $\checkmark$ & \textbf{98.88} & \textbf{3.73} & \textbf{3.81} & \textbf{98.39} & \textbf{99.16}\\
\midrule
\multirow{4}{*}{CIFAR-100}& \multirow{4}{*}{LSUNr} &- & - & - & 98.68 & 5.54 & 5.17 & 98.46 & 98.87\\
&&$\checkmark$ & - & - & 99.26 & 3.30 & 3.71 & 99.13 & 99.37 \\
&&$\checkmark$ & $\checkmark$ & - & 99.38 & 2.29 & 2.86 & 99.21 & 99.50 \\
&&$\checkmark$ & $\checkmark$ & $\checkmark$ & \textbf{99.70} & \textbf{1.09} & \textbf{1.77} & \textbf{99.61} & \textbf{99.76}\\
\bottomrule
\end{tabular}}
\label{tab:aba}
\end{table}

\subsection{Sensitivity of Hyperparameters}

We conduct numerous experiments on the hyperparameters appearing in Section~\ref{sec:method} to analyze their effect on TSL further.

\subsubsection{Analysis of $K$}

\begin{figure}[htbp]
\centering

\subfigure[CIFAR-10]{

\centering
\includegraphics[width=0.4\textwidth]{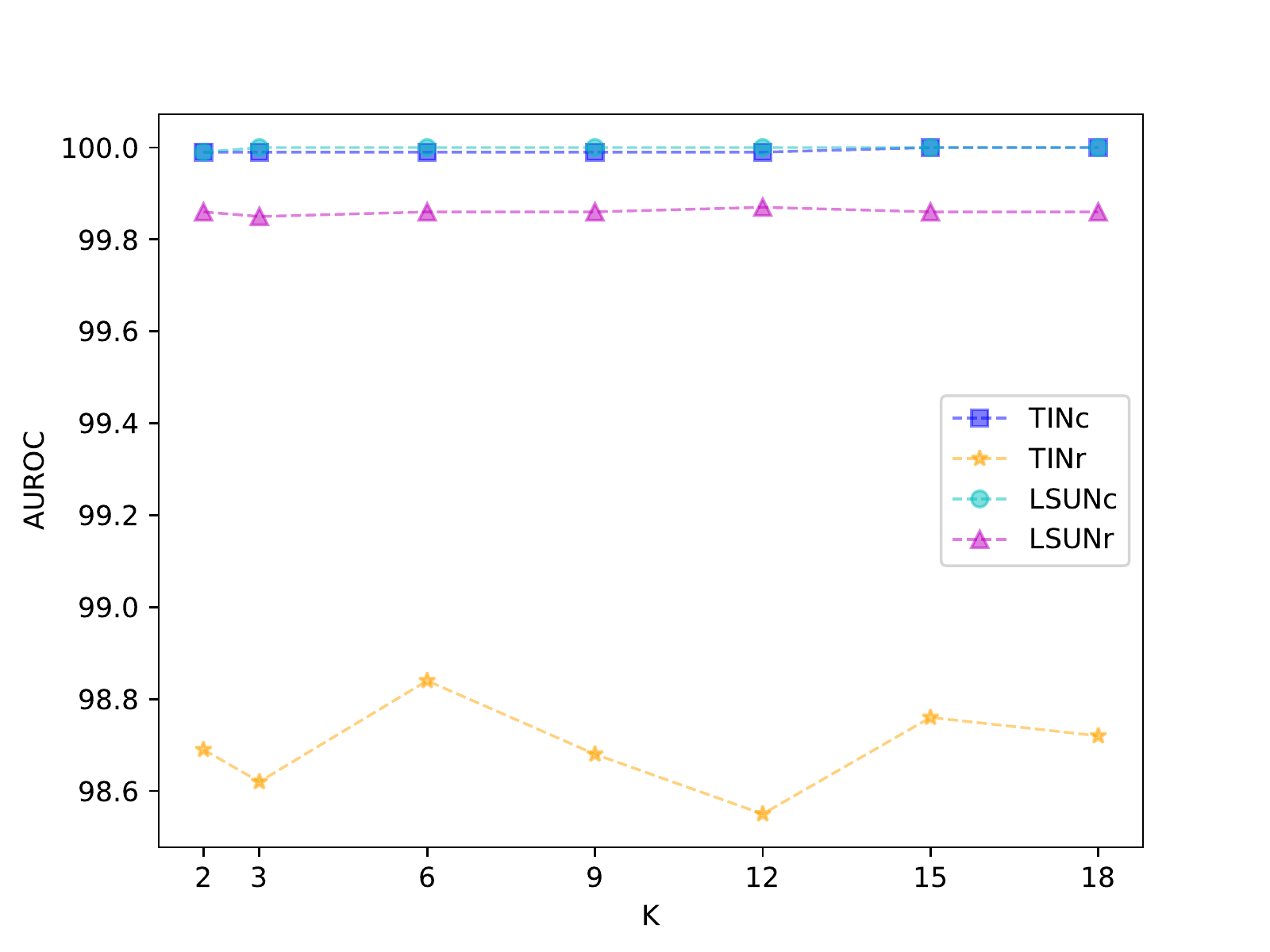}

}%
\subfigure[CIFAR-100]{

\centering
\includegraphics[width=0.4\textwidth]{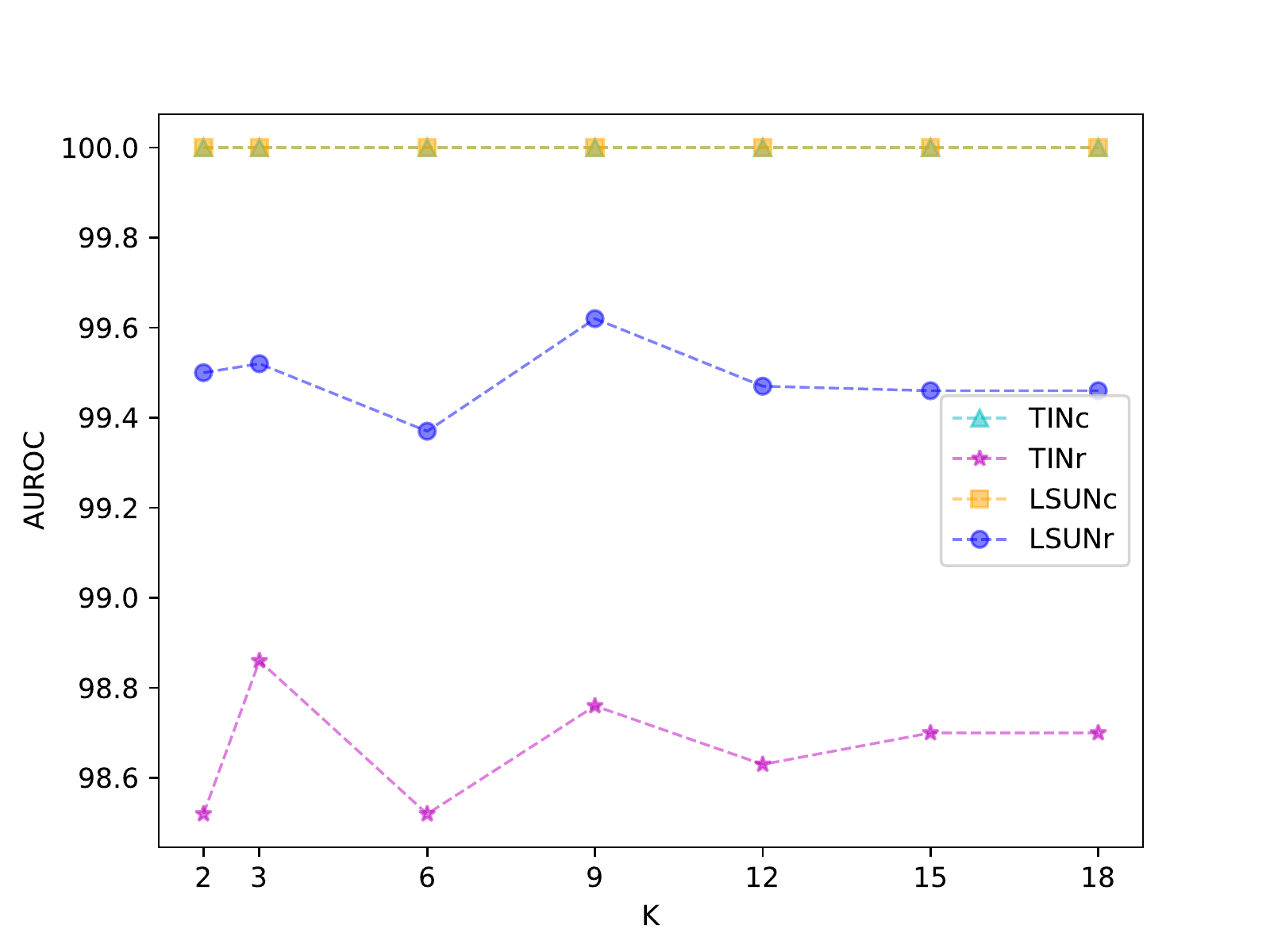}

}%
\centering
\caption{An analysis of AUROC at different $K$. (a) The variation of AUROC when the ID dataset is CIFAR-10. (b)The variation of AUROC when the ID dataset is CIFAR-100.
}
\label{K}
\end{figure}

$K$ is the number of neighbors for the KNN algorithm when mining positive pairs. We perform experiments to compare the performance of TSL on different ID and OOD dataset combinations for different $K$. From Fig.~\ref{K}, we can observe that TSL is insensitive to $K$ from 2 to 18.

\subsubsection{Analysis of M}

\begin{figure}[h]
\centering

\subfigure[$M$]{
\label{M}
\centering
\includegraphics[width=0.4\textwidth]{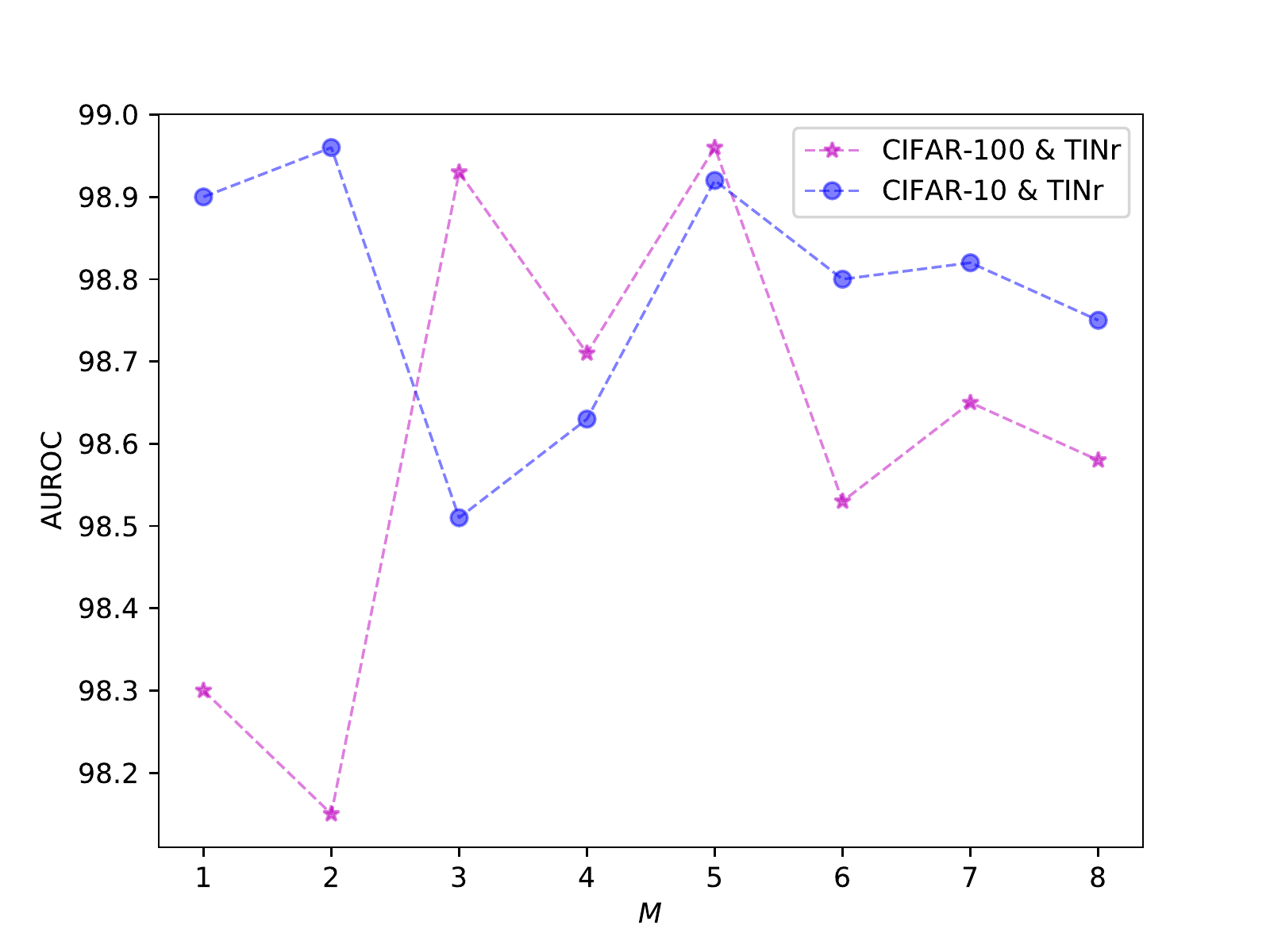}
%
}%
\subfigure[ $\lambda_1$]{
\label{lambda_1}
\centering
\includegraphics[width=0.4\textwidth]{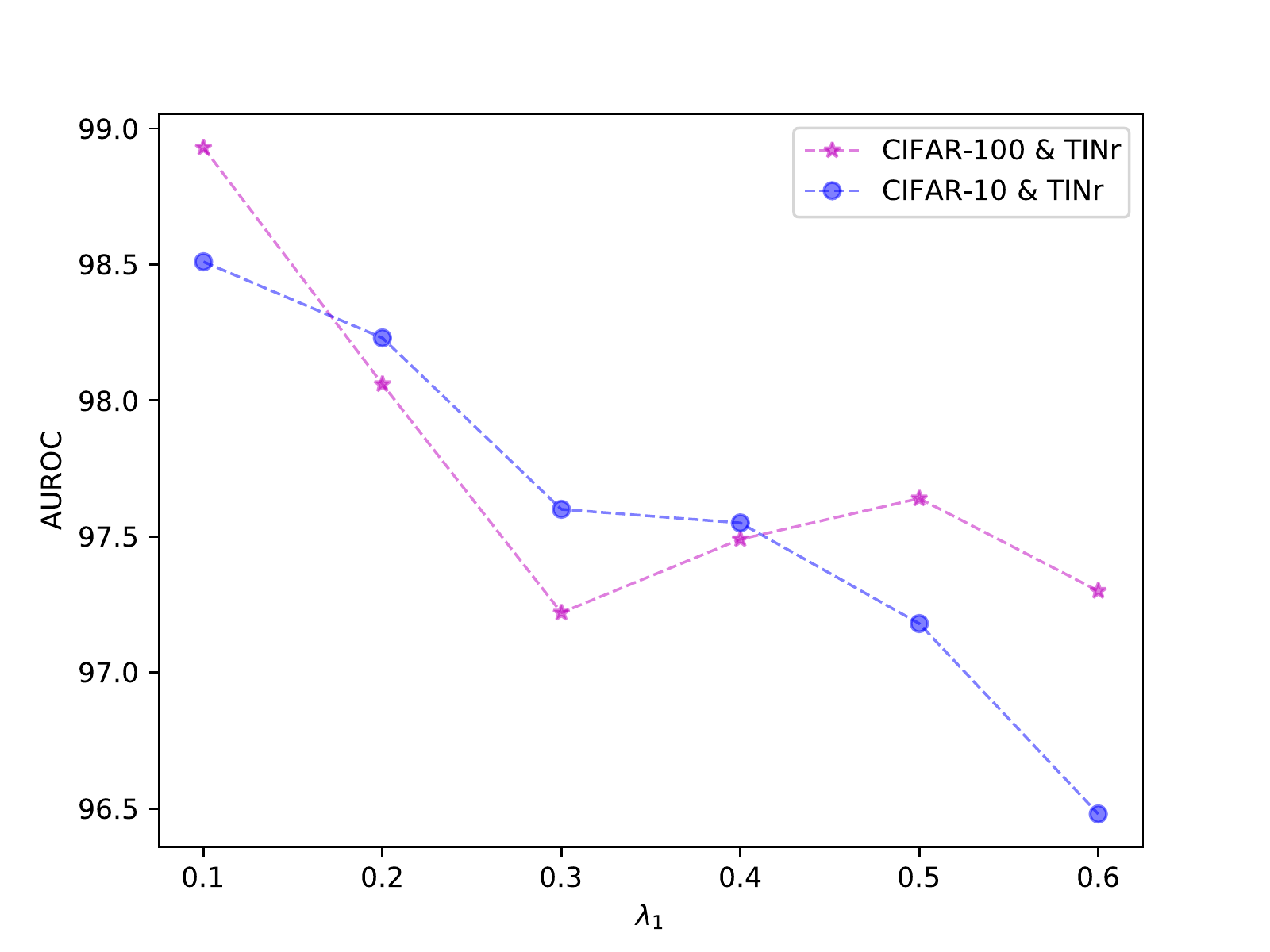}
}%
\centering
\caption{The analysis of different hyperparameters. (a) Variation of TSL's AUROC on two dataset combinations as the $M$ of $\mathcal{L}_{f}$ changes. (b) Variation of TSL's AUROC on two dataset combinations as the $\lambda_1$ of $\mathcal{L}_{a}$ changes. }
\label{aaaaa}
\end{figure}

$M$ is a coefficient responsible for regulating the distance between negative pairs when topological structure learning. Fig.~\ref{M} shows that TSL is insensitive to margin when $M$ is larger than three.

\subsubsection{Analysis of $\lambda_1$, $\lambda_2$, and $\lambda_3$}

\begin{table}[h]

\caption{An analysis of AUROC on CIFAR-100 \& TINr at different value of $\lambda_2$ .}
\centering
\label{l_1}
\scalebox{0.65}{
\begin{tabular}{c|cccccc}
\toprule
$\lambda_2$    & 0.1   & 0.3   & 0.5   & 0.7   & 0.9   & 1.0      \\ \midrule
AUROC $\uparrow$ & 97.75 & 98.43 & \textbf{98.93} & 98.68 & 98.98 & 98.61  \\ \midrule
$\lambda_2$  & 1.5   & 2.0   & 2.5   & 3.0   & 5.0   & 6.0\\\midrule
AUROC $\uparrow$ & 98.97 & 98.87 & 99.04 & \textbf{99.13} & 98.87 & 98.92\\
\bottomrule
\end{tabular}}
\end{table}

TSL has hyperparameters $\lambda_1$, $\lambda_2$, and $\lambda_3$. $\lambda_1$ regulates the closeness of the instances from the same class in $D_{L}$ for topological skeleton maintenance with ID labeled data. $\lambda_2$ and $\lambda_3$ are critical hyperparameters in topological skeleton extension for close positive pairs and loose positive pairs, respectively. We conduct numerous experiments to analyze them. First, Fig.~\ref{aaaaa} (b) shows that AUROC basically shows a decreasing trend as $\lambda_1$ increases, and TSL performs best when $\lambda_1$ takes the value of 0.1. So a smaller $\lambda_1$ helps improve TSL's performance. Second, from Table~\ref{l_1}, we observe that AUROC tends to level off after $\lambda_2>0.5$. Third, Table~\ref{l_2} shows how AUROC changes with different value of $\lambda_3$ on CIFAR-100 \& TINr. We observe that as $\lambda_3$ increases, AUROC first rises and then falls, and AUROC is maximum when $\lambda_3$ is 6.

\begin{table}[h]

\caption{An analysis of AUROC on CIFAR-100 \& TINr at different value of $\lambda_3$.}
\centering
\label{l_2}
\scalebox{0.65}{\begin{tabular}{c|ccccc}
\toprule
$\lambda_3$   & 1     & 2     & 4     & 6     & 8    \\ \midrule
AUROC $\uparrow$ & 98.50 & 98.45 & 98.71 & \textbf{98.93} & 98.75 \\ \midrule
$\lambda_3$   & 10    & 12    & 14    & 17    & 20    \\ \midrule
AUROC $\uparrow$ & 98.76 & 98.54 & 98.56 & 98.37 & 98.38 \\\bottomrule
\end{tabular}}
\end{table}

\subsubsection{Analysis of $\beta$}
As can be seen from Table~\ref{beta}, TSL maintains good robustness when $\beta$ goes from 1 to 4,500 on all CIFAR-100 \& TINr. However, after $\beta$ is greater than 4,500, the performance of TSL starts to drop sharply as $\beta$ increases.

\begin{table}[t]
\caption{An analysis of AUROC on CIFAR-100 \& TINr at different value of $\beta$.}
\centering
\label{beta}
\scalebox{0.65}{\begin{tabular}{c|cccccc}
\toprule
$\beta$  & 0 & 1         & 100   & 500   & 1000  \\ \midrule
AUROC $\uparrow$ &98.06  &\textbf{98.93}  & 98.85 & 98.74 & 98.78 \\ \midrule
$\beta$  & 2000  & 4000  & 4500  & 4700  & 4800  \\ \midrule
AUROC $\uparrow$ & 98.62 & 98.86 & 98.89 & 98.53 & 98.24 \\ \bottomrule
\end{tabular}}
\end{table}

\section{Conclusion}\label{sec:conclusion}
In this paper, we explored a new problem, weakly-supervised OOD detection, and analyzed why limited ID labeled instances brought challenges. To solve this new problem, we proposed a novel OOD detection method called TSL. TSL constructed the initial points for the topological space by extracting reliable features with SimCLR. In the initial topological space, TSL mined topological connections, including close positive pairs, loose positive pairs, and negative pairs, to construct several different kinds of edges with different credibility. Moreover, TSL reconstructs the topological structure in a new topological space with two steps, topological skeleton maintenance and topological skeleton extension, to increase the separability of ID and OOD instances. Empirical studies showed that TSL achieved stable performance gain across different OOD datasets and different tasks, which verified the validity and robustness of TSL. Beyond this work, this problem is more relevant to real-world applications, so it is worth studying in the future.

\section*{Acknowledgment}
This work is supported by the National Natural Science Foundation of China (62176139, 61876098), the Major Basic Research Project of Natural Science Foundation of Shandong Province (ZR2021ZD15).

\bibliography{mybibfile}

\begin{thebibliography}{51}
\expandafter\ifx\csname natexlab\endcsname\relax\def\natexlab#1{#1}\fi
\providecommand{\url}[1]{\texttt{#1}}
\providecommand{\href}[2]{#2}
\providecommand{\path}[1]{#1}
\providecommand{\DOIprefix}{doi:}
\providecommand{\ArXivprefix}{arXiv:}
\providecommand{\URLprefix}{URL: }
\providecommand{\Pubmedprefix}{pmid:}
\providecommand{\doi}[1]{\href{http://dx.doi.org/#1}{\path{#1}}}
\providecommand{\Pubmed}[1]{\href{pmid:#1}{\path{#1}}}
\providecommand{\bibinfo}[2]{#2}
\ifx\xfnm\relax \def\xfnm[#1]{\unskip,\space#1}\fi
\bibitem[{Bustos et~al.(2020)Bustos, Pertusa, Salinas and de~la
  Iglesia-Vay{\'a}}]{bustos2020padchest}
\bibinfo{author}{Bustos, A.}, \bibinfo{author}{Pertusa, A.},
  \bibinfo{author}{Salinas, J.M.}, \bibinfo{author}{de~la Iglesia-Vay{\'a},
  M.}, \bibinfo{year}{2020}.
\newblock \bibinfo{title}{Padchest: A large chest x-ray image dataset with
  multi-label annotated reports}.
\newblock \bibinfo{journal}{Medical image analysis} \bibinfo{volume}{66},
  \bibinfo{pages}{101797}.
\bibitem[{Canziani et~al.(2016)Canziani, Paszke and
  Culurciello}]{canziani2016analysis}
\bibinfo{author}{Canziani, A.}, \bibinfo{author}{Paszke, A.},
  \bibinfo{author}{Culurciello, E.}, \bibinfo{year}{2016}.
\newblock \bibinfo{title}{An analysis of deep neural network models for
  practical applications}.
\newblock \bibinfo{journal}{arXiv preprint arXiv:1605.07678} .
\bibitem[{Carbonneau et~al.(2018)Carbonneau, Cheplygina, Granger and
  Gagnon}]{carbonneau2018multiple}
\bibinfo{author}{Carbonneau, M.A.}, \bibinfo{author}{Cheplygina, V.},
  \bibinfo{author}{Granger, E.}, \bibinfo{author}{Gagnon, G.},
  \bibinfo{year}{2018}.
\newblock \bibinfo{title}{Multiple instance learning: A survey of problem
  characteristics and applications}.
\newblock \bibinfo{journal}{Pattern Recognition} \bibinfo{volume}{77},
  \bibinfo{pages}{329--353}.
\bibitem[{Chapelle et~al.(2009)Chapelle, Scholkopf and Zien}]{chapelle2009semi}
\bibinfo{author}{Chapelle, O.}, \bibinfo{author}{Scholkopf, B.},
  \bibinfo{author}{Zien, A.}, \bibinfo{year}{2009}.
\newblock \bibinfo{title}{Semi-supervised learning (chapelle, o. et al., eds.;
  2006)[book reviews]}.
\newblock \bibinfo{journal}{IEEE Transactions on Neural Networks}
  \bibinfo{volume}{20}, \bibinfo{pages}{542--542}.
\bibitem[{Chen et~al.(2021)Chen, Li, Wu, Liang and Jha}]{chen2021atom}
\bibinfo{author}{Chen, J.}, \bibinfo{author}{Li, Y.}, \bibinfo{author}{Wu, X.},
  \bibinfo{author}{Liang, Y.}, \bibinfo{author}{Jha, S.}, \bibinfo{year}{2021}.
\newblock \bibinfo{title}{Atom: Robustifying out-of-distribution detection
  using outlier mining}, in: \bibinfo{booktitle}{Joint European Conference on
  Machine Learning and Knowledge Discovery in Databases},
  \bibinfo{organization}{Springer}. pp. \bibinfo{pages}{430--445}.
\bibitem[{Chen et~al.(2020)Chen, Kornblith, Norouzi and
  Hinton}]{chen2020simple}
\bibinfo{author}{Chen, T.}, \bibinfo{author}{Kornblith, S.},
  \bibinfo{author}{Norouzi, M.}, \bibinfo{author}{Hinton, G.},
  \bibinfo{year}{2020}.
\newblock \bibinfo{title}{A simple framework for contrastive learning of visual
  representations}, in: \bibinfo{booktitle}{International conference on machine
  learning}, \bibinfo{organization}{PMLR}. pp. \bibinfo{pages}{1597--1607}.
\bibitem[{Chen et~al.(2017)Chen, Sun, Shi and Hong}]{chen2017sampling}
\bibinfo{author}{Chen, T.}, \bibinfo{author}{Sun, Y.}, \bibinfo{author}{Shi,
  Y.}, \bibinfo{author}{Hong, L.}, \bibinfo{year}{2017}.
\newblock \bibinfo{title}{On sampling strategies for neural network-based
  collaborative filtering}, in: \bibinfo{booktitle}{Proceedings of the 23rd ACM
  SIGKDD International Conference on Knowledge Discovery and Data Mining}, pp.
  \bibinfo{pages}{767--776}.
\bibitem[{Deng et~al.(2009)Deng, Dong, Socher, Li, Li and
  Fei-Fei}]{deng2009imagenet}
\bibinfo{author}{Deng, J.}, \bibinfo{author}{Dong, W.},
  \bibinfo{author}{Socher, R.}, \bibinfo{author}{Li, L.J.},
  \bibinfo{author}{Li, K.}, \bibinfo{author}{Fei-Fei, L.},
  \bibinfo{year}{2009}.
\newblock \bibinfo{title}{Imagenet: A large-scale hierarchical image database},
  in: \bibinfo{booktitle}{2009 IEEE conference on computer vision and pattern
  recognition}, \bibinfo{organization}{Ieee}. pp. \bibinfo{pages}{248--255}.
\bibitem[{Dong et~al.(2017)Dong, Pan, Zhang and Xu}]{dong2017learning}
\bibinfo{author}{Dong, Y.}, \bibinfo{author}{Pan, Y.}, \bibinfo{author}{Zhang,
  J.}, \bibinfo{author}{Xu, W.}, \bibinfo{year}{2017}.
\newblock \bibinfo{title}{Learning to read chest x-ray images from 16000+
  examples using cnn}, in: \bibinfo{booktitle}{2017 IEEE/ACM International
  Conference on Connected Health: Applications, Systems and Engineering
  Technologies (CHASE)}, \bibinfo{organization}{IEEE}. pp.
  \bibinfo{pages}{51--57}.
\bibitem[{Du et~al.(2022)Du, Wang, Cai and Li}]{du2022vos}
\bibinfo{author}{Du, X.}, \bibinfo{author}{Wang, Z.}, \bibinfo{author}{Cai,
  M.}, \bibinfo{author}{Li, Y.}, \bibinfo{year}{2022}.
\newblock \bibinfo{title}{Vos: Learning what you don't know by virtual outlier
  synthesis}.
\newblock \bibinfo{journal}{arXiv preprint arXiv:2202.01197} .
\bibitem[{Fr{\'e}nay and Verleysen(2013)}]{frenay2013classification}
\bibinfo{author}{Fr{\'e}nay, B.}, \bibinfo{author}{Verleysen, M.},
  \bibinfo{year}{2013}.
\newblock \bibinfo{title}{Classification in the presence of label noise: a
  survey}.
\newblock \bibinfo{journal}{IEEE transactions on neural networks and learning
  systems} \bibinfo{volume}{25}, \bibinfo{pages}{845--869}.
\bibitem[{Han et~al.(2021)Han, He, Li, Wei, Wang and Yin}]{han2021semi}
\bibinfo{author}{Han, Z.}, \bibinfo{author}{He, R.}, \bibinfo{author}{Li, T.},
  \bibinfo{author}{Wei, B.}, \bibinfo{author}{Wang, J.}, \bibinfo{author}{Yin,
  Y.}, \bibinfo{year}{2021}.
\newblock \bibinfo{title}{Semi-supervised screening of covid-19 from positive
  and unlabeled data with constraint non-negative risk estimator}, in:
  \bibinfo{booktitle}{International Conference on Information Processing in
  Medical Imaging}, \bibinfo{organization}{Springer}. pp.
  \bibinfo{pages}{611--623}.
\bibitem[{Hao et~al.(2020)Hao, Liu, Zhang, Zhu, Chen, Jiang and
  Ngo}]{hao2020person}
\bibinfo{author}{Hao, Y.}, \bibinfo{author}{Liu, Z.N.}, \bibinfo{author}{Zhang,
  H.}, \bibinfo{author}{Zhu, B.}, \bibinfo{author}{Chen, J.},
  \bibinfo{author}{Jiang, Y.G.}, \bibinfo{author}{Ngo, C.W.},
  \bibinfo{year}{2020}.
\newblock \bibinfo{title}{Person-level action recognition in complex events via
  tsd-tsm networks}, in: \bibinfo{booktitle}{Proceedings of the 28th ACM
  International Conference on Multimedia}, pp. \bibinfo{pages}{4699--4702}.
\bibitem[{HaoChen et~al.(2021)HaoChen, Wei, Gaidon and
  Ma}]{haochen2021provable}
\bibinfo{author}{HaoChen, J.Z.}, \bibinfo{author}{Wei, C.},
  \bibinfo{author}{Gaidon, A.}, \bibinfo{author}{Ma, T.}, \bibinfo{year}{2021}.
\newblock \bibinfo{title}{Provable guarantees for self-supervised deep learning
  with spectral contrastive loss}.
\newblock \bibinfo{journal}{Advances in Neural Information Processing Systems}
  \bibinfo{volume}{34}.
\bibitem[{He et~al.(2022)He, Han and Yin}]{he2022towards}
\bibinfo{author}{He, R.}, \bibinfo{author}{Han, Z.}, \bibinfo{author}{Yin, Y.},
  \bibinfo{year}{2022}.
\newblock \bibinfo{title}{Towards safe and robust weakly-supervised anomaly
  detection under subpopulation shift}.
\newblock \bibinfo{journal}{Knowledge-Based Systems} , \bibinfo{pages}{109088}.
\bibitem[{He et~al.(2021)He, Han, Zhang, He, Nie and Yin}]{he2021robust}
\bibinfo{author}{He, R.}, \bibinfo{author}{Han, Z.}, \bibinfo{author}{Zhang,
  Y.}, \bibinfo{author}{He, X.}, \bibinfo{author}{Nie, X.},
  \bibinfo{author}{Yin, Y.}, \bibinfo{year}{2021}.
\newblock \bibinfo{title}{Robust anomaly detection from partially observed
  anomalies with augmented classes}, in: \bibinfo{booktitle}{CAAI International
  Conference on Artificial Intelligence}, \bibinfo{organization}{Springer}. pp.
  \bibinfo{pages}{347--358}.
\bibitem[{Hell et~al.(2021)Hell, Hinz, Liu, Goyal, Pei, Lytvynenko, Knoll and
  Yiqiang}]{hell2021monitoring}
\bibinfo{author}{Hell, F.}, \bibinfo{author}{Hinz, G.}, \bibinfo{author}{Liu,
  F.}, \bibinfo{author}{Goyal, S.}, \bibinfo{author}{Pei, K.},
  \bibinfo{author}{Lytvynenko, T.}, \bibinfo{author}{Knoll, A.},
  \bibinfo{author}{Yiqiang, C.}, \bibinfo{year}{2021}.
\newblock \bibinfo{title}{Monitoring perception reliability in autonomous
  driving: Distributional shift detection for estimating the impact of input
  data on prediction accuracy}, in: \bibinfo{booktitle}{Computer Science in
  Cars Symposium}, pp. \bibinfo{pages}{1--9}.
\bibitem[{Hendrycks and Gimpel(2016)}]{hendrycks2016baseline}
\bibinfo{author}{Hendrycks, D.}, \bibinfo{author}{Gimpel, K.},
  \bibinfo{year}{2016}.
\newblock \bibinfo{title}{A baseline for detecting misclassified and
  out-of-distribution examples in neural networks}.
\newblock \bibinfo{journal}{arXiv preprint arXiv:1610.02136} .
\bibitem[{Hendrycks et~al.(2018)Hendrycks, Mazeika and
  Dietterich}]{hendrycks2018deep}
\bibinfo{author}{Hendrycks, D.}, \bibinfo{author}{Mazeika, M.},
  \bibinfo{author}{Dietterich, T.}, \bibinfo{year}{2018}.
\newblock \bibinfo{title}{Deep anomaly detection with outlier exposure}.
\newblock \bibinfo{journal}{arXiv preprint arXiv:1812.04606} .
\bibitem[{Huang et~al.(2017)Huang, Liu, Van Der~Maaten and
  Weinberger}]{huang2017densely}
\bibinfo{author}{Huang, G.}, \bibinfo{author}{Liu, Z.}, \bibinfo{author}{Van
  Der~Maaten, L.}, \bibinfo{author}{Weinberger, K.Q.}, \bibinfo{year}{2017}.
\newblock \bibinfo{title}{Densely connected convolutional networks}, in:
  \bibinfo{booktitle}{Proceedings of the IEEE conference on computer vision and
  pattern recognition}, pp. \bibinfo{pages}{4700--4708}.
\bibitem[{Jeong and Kim(2020)}]{jeong2020ood}
\bibinfo{author}{Jeong, T.}, \bibinfo{author}{Kim, H.}, \bibinfo{year}{2020}.
\newblock \bibinfo{title}{Ood-maml: Meta-learning for few-shot
  out-of-distribution detection and classification}.
\newblock \bibinfo{journal}{Advances in Neural Information Processing Systems}
  \bibinfo{volume}{33}, \bibinfo{pages}{3907--3916}.
\bibitem[{Katz-Samuels et~al.(2022)Katz-Samuels, Nakhleh, Nowak and
  Li}]{katz2022training}
\bibinfo{author}{Katz-Samuels, J.}, \bibinfo{author}{Nakhleh, J.},
  \bibinfo{author}{Nowak, R.}, \bibinfo{author}{Li, Y.}, \bibinfo{year}{2022}.
\newblock \bibinfo{title}{Training ood detectors in their natural habitats}.
\newblock \bibinfo{journal}{arXiv preprint arXiv:2202.03299} .
\bibitem[{Krizhevsky et~al.(2009)Krizhevsky, Hinton
  et~al.}]{krizhevsky2009learning}
\bibinfo{author}{Krizhevsky, A.}, \bibinfo{author}{Hinton, G.}, et~al.,
  \bibinfo{year}{2009}.
\newblock \bibinfo{title}{Learning multiple layers of features from tiny
  images} .
\bibitem[{Lee et~al.(2018)Lee, Lee, Lee and Shin}]{lee2018simple}
\bibinfo{author}{Lee, K.}, \bibinfo{author}{Lee, K.}, \bibinfo{author}{Lee,
  H.}, \bibinfo{author}{Shin, J.}, \bibinfo{year}{2018}.
\newblock \bibinfo{title}{A simple unified framework for detecting
  out-of-distribution samples and adversarial attacks}.
\newblock \bibinfo{journal}{Advances in neural information processing systems}
  \bibinfo{volume}{31}.
\bibitem[{Lee and Liu(2003)}]{lee2003learning}
\bibinfo{author}{Lee, W.S.}, \bibinfo{author}{Liu, B.}, \bibinfo{year}{2003}.
\newblock \bibinfo{title}{Learning with positive and unlabeled examples using
  weighted logistic regression}, in: \bibinfo{booktitle}{ICML}, pp.
  \bibinfo{pages}{448--455}.
\bibitem[{Li et~al.(2019)Li, Guo and Zhou}]{li2019towards}
\bibinfo{author}{Li, Y.F.}, \bibinfo{author}{Guo, L.Z.}, \bibinfo{author}{Zhou,
  Z.H.}, \bibinfo{year}{2019}.
\newblock \bibinfo{title}{Towards safe weakly supervised learning}.
\newblock \bibinfo{journal}{IEEE transactions on pattern analysis and machine
  intelligence} \bibinfo{volume}{43}, \bibinfo{pages}{334--346}.
\bibitem[{Liang et~al.(2017)Liang, Li and Srikant}]{liang2017enhancing}
\bibinfo{author}{Liang, S.}, \bibinfo{author}{Li, Y.},
  \bibinfo{author}{Srikant, R.}, \bibinfo{year}{2017}.
\newblock \bibinfo{title}{Enhancing the reliability of out-of-distribution
  image detection in neural networks}.
\newblock \bibinfo{journal}{arXiv preprint arXiv:1706.02690} .
\bibitem[{Liu et~al.(2003)Liu, Dai, Li, Lee and Yu}]{liu2003building}
\bibinfo{author}{Liu, B.}, \bibinfo{author}{Dai, Y.}, \bibinfo{author}{Li, X.},
  \bibinfo{author}{Lee, W.S.}, \bibinfo{author}{Yu, P.S.},
  \bibinfo{year}{2003}.
\newblock \bibinfo{title}{Building text classifiers using positive and
  unlabeled examples}, in: \bibinfo{booktitle}{Third IEEE international
  conference on data mining}, \bibinfo{organization}{IEEE}. pp.
  \bibinfo{pages}{179--186}.
\bibitem[{Liu et~al.(2002)Liu, Lee, Yu and Li}]{liu2002partially}
\bibinfo{author}{Liu, B.}, \bibinfo{author}{Lee, W.S.}, \bibinfo{author}{Yu,
  P.S.}, \bibinfo{author}{Li, X.}, \bibinfo{year}{2002}.
\newblock \bibinfo{title}{Partially supervised classification of text
  documents}, in: \bibinfo{booktitle}{ICML}, \bibinfo{organization}{Sydney,
  NSW}. pp. \bibinfo{pages}{387--394}.
\bibitem[{Liu et~al.(2020)Liu, Wang, Owens and Li}]{liu2020energy}
\bibinfo{author}{Liu, W.}, \bibinfo{author}{Wang, X.}, \bibinfo{author}{Owens,
  J.D.}, \bibinfo{author}{Li, Y.}, \bibinfo{year}{2020}.
\newblock \bibinfo{title}{Energy-based out-of-distribution detection}.
\newblock \bibinfo{journal}{arXiv preprint arXiv:2010.03759} .
\bibitem[{McLachlan(1999)}]{mclachlan1999mahalanobis}
\bibinfo{author}{McLachlan, G.J.}, \bibinfo{year}{1999}.
\newblock \bibinfo{title}{Mahalanobis distance}.
\newblock \bibinfo{journal}{Resonance} \bibinfo{volume}{4},
  \bibinfo{pages}{20--26}.
\bibitem[{Ming et~al.(2022)Ming, Fan and Li}]{ming2022poem}
\bibinfo{author}{Ming, Y.}, \bibinfo{author}{Fan, Y.}, \bibinfo{author}{Li,
  Y.}, \bibinfo{year}{2022}.
\newblock \bibinfo{title}{Poem: Out-of-distribution detection with posterior
  sampling}, in: \bibinfo{booktitle}{International Conference on Machine
  Learning}, \bibinfo{organization}{PMLR}. pp. \bibinfo{pages}{15650--15665}.
\bibitem[{Morteza and Li(2021)}]{morteza2021provable}
\bibinfo{author}{Morteza, P.}, \bibinfo{author}{Li, Y.}, \bibinfo{year}{2021}.
\newblock \bibinfo{title}{Provable guarantees for understanding
  out-of-distribution detection}.
\newblock \bibinfo{journal}{arXiv preprint arXiv:2112.00787} .
\bibitem[{Nguyen et~al.(2015)Nguyen, Yosinski and Clune}]{nguyen2015deep}
\bibinfo{author}{Nguyen, A.}, \bibinfo{author}{Yosinski, J.},
  \bibinfo{author}{Clune, J.}, \bibinfo{year}{2015}.
\newblock \bibinfo{title}{Deep neural networks are easily fooled: High
  confidence predictions for unrecognizable images}, in:
  \bibinfo{booktitle}{Proceedings of the IEEE conference on computer vision and
  pattern recognition}, pp. \bibinfo{pages}{427--436}.
\bibitem[{Nguyen et~al.(2017)Nguyen, Maclagan, Nguyen, Nguyen, Flemons,
  Andrews, Ritchie and Phung}]{nguyen2017animal}
\bibinfo{author}{Nguyen, H.}, \bibinfo{author}{Maclagan, S.J.},
  \bibinfo{author}{Nguyen, T.D.}, \bibinfo{author}{Nguyen, T.},
  \bibinfo{author}{Flemons, P.}, \bibinfo{author}{Andrews, K.},
  \bibinfo{author}{Ritchie, E.G.}, \bibinfo{author}{Phung, D.},
  \bibinfo{year}{2017}.
\newblock \bibinfo{title}{Animal recognition and identification with deep
  convolutional neural networks for automated wildlife monitoring}, in:
  \bibinfo{booktitle}{2017 IEEE international conference on data science and
  advanced Analytics (DSAA)}, \bibinfo{organization}{IEEE}. pp.
  \bibinfo{pages}{40--49}.
\bibitem[{Nitsch et~al.(2021)Nitsch, Itkina, Senanayake, Nieto, Schmidt,
  Siegwart, Kochenderfer and Cadena}]{nitsch2021out}
\bibinfo{author}{Nitsch, J.}, \bibinfo{author}{Itkina, M.},
  \bibinfo{author}{Senanayake, R.}, \bibinfo{author}{Nieto, J.},
  \bibinfo{author}{Schmidt, M.}, \bibinfo{author}{Siegwart, R.},
  \bibinfo{author}{Kochenderfer, M.J.}, \bibinfo{author}{Cadena, C.},
  \bibinfo{year}{2021}.
\newblock \bibinfo{title}{Out-of-distribution detection for automotive
  perception}, in: \bibinfo{booktitle}{2021 IEEE International Intelligent
  Transportation Systems Conference (ITSC)}, \bibinfo{organization}{IEEE}. pp.
  \bibinfo{pages}{2938--2943}.
\bibitem[{Ren et~al.(2019)Ren, Liu, Fertig, Snoek, Poplin, DePristo, Dillon and
  Lakshminarayanan}]{ren2019likelihood}
\bibinfo{author}{Ren, J.}, \bibinfo{author}{Liu, P.J.},
  \bibinfo{author}{Fertig, E.}, \bibinfo{author}{Snoek, J.},
  \bibinfo{author}{Poplin, R.}, \bibinfo{author}{DePristo, M.A.},
  \bibinfo{author}{Dillon, J.V.}, \bibinfo{author}{Lakshminarayanan, B.},
  \bibinfo{year}{2019}.
\newblock \bibinfo{title}{Likelihood ratios for out-of-distribution detection}.
\newblock \bibinfo{journal}{arXiv preprint arXiv:1906.02845} .
\bibitem[{Shafaei et~al.(2018)Shafaei, Schmidt and Little}]{shafaei2018less}
\bibinfo{author}{Shafaei, A.}, \bibinfo{author}{Schmidt, M.},
  \bibinfo{author}{Little, J.J.}, \bibinfo{year}{2018}.
\newblock \bibinfo{title}{A less biased evaluation of out-of-distribution
  sample detectors}.
\newblock \bibinfo{journal}{arXiv preprint arXiv:1809.04729} .
\bibitem[{Sun et~al.(2021)Sun, Yang, Zhang, Liu, Halappanavar, Fan and
  Cao}]{sun2021gradient}
\bibinfo{author}{Sun, J.}, \bibinfo{author}{Yang, L.}, \bibinfo{author}{Zhang,
  J.}, \bibinfo{author}{Liu, F.}, \bibinfo{author}{Halappanavar, M.},
  \bibinfo{author}{Fan, D.}, \bibinfo{author}{Cao, Y.}, \bibinfo{year}{2021}.
\newblock \bibinfo{title}{Gradient-based novelty detection boosted by
  self-supervised binary classification}.
\newblock \bibinfo{journal}{arXiv preprint arXiv:2112.09815} .
\bibitem[{Tack et~al.(2020)Tack, Mo, Jeong and Shin}]{tack2020csi}
\bibinfo{author}{Tack, J.}, \bibinfo{author}{Mo, S.}, \bibinfo{author}{Jeong,
  J.}, \bibinfo{author}{Shin, J.}, \bibinfo{year}{2020}.
\newblock \bibinfo{title}{Csi: Novelty detection via contrastive learning on
  distributionally shifted instances}.
\newblock \bibinfo{journal}{Advances in neural information processing systems}
  \bibinfo{volume}{33}, \bibinfo{pages}{11839--11852}.
\bibitem[{Uwimana and Senanayake(2021)}]{uwimana2021out}
\bibinfo{author}{Uwimana, A.}, \bibinfo{author}{Senanayake, R.},
  \bibinfo{year}{2021}.
\newblock \bibinfo{title}{Out of distribution detection and adversarial attacks
  on deep neural networks for robust medical image analysis}.
\newblock \bibinfo{journal}{arXiv preprint arXiv:2107.04882} .
\bibitem[{Wang et~al.(2021)Wang, Liu, Bocchieri and Li}]{wang2021can}
\bibinfo{author}{Wang, H.}, \bibinfo{author}{Liu, W.},
  \bibinfo{author}{Bocchieri, A.}, \bibinfo{author}{Li, Y.},
  \bibinfo{year}{2021}.
\newblock \bibinfo{title}{Can multi-label classification networks know what
  they don’t know?}
\newblock \bibinfo{journal}{Advances in Neural Information Processing Systems}
  \bibinfo{volume}{34}.
\bibitem[{Xiao et~al.(2020)Xiao, Yan and Amit}]{xiao2020likelihood}
\bibinfo{author}{Xiao, Z.}, \bibinfo{author}{Yan, Q.}, \bibinfo{author}{Amit,
  Y.}, \bibinfo{year}{2020}.
\newblock \bibinfo{title}{Likelihood regret: An out-of-distribution detection
  score for variational auto-encoder}.
\newblock \bibinfo{journal}{arXiv preprint arXiv:2003.02977} .
\bibitem[{Yang et~al.(2021a)Yang, Wang, Feng, Yan, Zheng, Zhang and
  Liu}]{yang2021semantically}
\bibinfo{author}{Yang, J.}, \bibinfo{author}{Wang, H.}, \bibinfo{author}{Feng,
  L.}, \bibinfo{author}{Yan, X.}, \bibinfo{author}{Zheng, H.},
  \bibinfo{author}{Zhang, W.}, \bibinfo{author}{Liu, Z.},
  \bibinfo{year}{2021}a.
\newblock \bibinfo{title}{Semantically coherent out-of-distribution detection},
  in: \bibinfo{booktitle}{Proceedings of the IEEE/CVF International Conference
  on Computer Vision}, pp. \bibinfo{pages}{8301--8309}.
\bibitem[{Yang et~al.(2021b)Yang, Zhou, Li and Liu}]{yang2021generalized}
\bibinfo{author}{Yang, J.}, \bibinfo{author}{Zhou, K.}, \bibinfo{author}{Li,
  Y.}, \bibinfo{author}{Liu, Z.}, \bibinfo{year}{2021}b.
\newblock \bibinfo{title}{Generalized out-of-distribution detection: A survey}.
\newblock \bibinfo{journal}{arXiv preprint arXiv:2110.11334} .
\bibitem[{Yu et~al.(2015)Yu, Seff, Zhang, Song, Funkhouser and
  Xiao}]{yu2015lsun}
\bibinfo{author}{Yu, F.}, \bibinfo{author}{Seff, A.}, \bibinfo{author}{Zhang,
  Y.}, \bibinfo{author}{Song, S.}, \bibinfo{author}{Funkhouser, T.},
  \bibinfo{author}{Xiao, J.}, \bibinfo{year}{2015}.
\newblock \bibinfo{title}{Lsun: Construction of a large-scale image dataset
  using deep learning with humans in the loop}.
\newblock \bibinfo{journal}{arXiv preprint arXiv:1506.03365} .
\bibitem[{Yu and Aizawa(2019)}]{yu2019unsupervised}
\bibinfo{author}{Yu, Q.}, \bibinfo{author}{Aizawa, K.}, \bibinfo{year}{2019}.
\newblock \bibinfo{title}{Unsupervised out-of-distribution detection by maximum
  classifier discrepancy}, in: \bibinfo{booktitle}{Proceedings of the IEEE/CVF
  International Conference on Computer Vision}, pp.
  \bibinfo{pages}{9518--9526}.
\bibitem[{Zhang et~al.(2018)Zhang, Li, Zhou, Li and Zhou}]{zhang2018anomaly}
\bibinfo{author}{Zhang, Y.L.}, \bibinfo{author}{Li, L.}, \bibinfo{author}{Zhou,
  J.}, \bibinfo{author}{Li, X.}, \bibinfo{author}{Zhou, Z.H.},
  \bibinfo{year}{2018}.
\newblock \bibinfo{title}{Anomaly detection with partially observed anomalies},
  in: \bibinfo{booktitle}{Companion Proceedings of the The Web Conference
  2018}, pp. \bibinfo{pages}{639--646}.
\bibitem[{Zhou et~al.(2021)Zhou, Guo, Cheng, Li and Pu}]{zhou2021step}
\bibinfo{author}{Zhou, Z.}, \bibinfo{author}{Guo, L.Z.},
  \bibinfo{author}{Cheng, Z.}, \bibinfo{author}{Li, Y.F.}, \bibinfo{author}{Pu,
  S.}, \bibinfo{year}{2021}.
\newblock \bibinfo{title}{Step: Out-of-distribution detection in the presence
  of limited in-distribution labeled data}.
\newblock \bibinfo{journal}{Advances in Neural Information Processing Systems}
  \bibinfo{volume}{34}.
\bibitem[{Zhou(2018)}]{zhou2018brief}
\bibinfo{author}{Zhou, Z.H.}, \bibinfo{year}{2018}.
\newblock \bibinfo{title}{A brief introduction to weakly supervised learning}.
\newblock \bibinfo{journal}{National science review} \bibinfo{volume}{5},
  \bibinfo{pages}{44--53}.
\bibitem[{Zisselman and Tamar(2020)}]{zisselman2020deep}
\bibinfo{author}{Zisselman, E.}, \bibinfo{author}{Tamar, A.},
  \bibinfo{year}{2020}.
\newblock \bibinfo{title}{Deep residual flow for out of distribution
  detection}, in: \bibinfo{booktitle}{Proceedings of the IEEE/CVF Conference on
  Computer Vision and Pattern Recognition}, pp. \bibinfo{pages}{13994--14003}.

\end{thebibliography}

\end{document}